\documentclass{article}

     \PassOptionsToPackage{numbers, compress}{natbib}

          \usepackage[final]{neurips_2019}

\usepackage[utf8]{inputenc} 
\usepackage[T1]{fontenc}    
\usepackage[breaklinks=true]{hyperref}     
\usepackage{url}            
\usepackage{booktabs}       
\usepackage{amsfonts}       
\usepackage{nicefrac}       
\usepackage{microtype}      
\usepackage{booktabs} 
\usepackage{subcaption}
\usepackage{listings}
\usepackage{graphicx}
\usepackage{color}
\usepackage{longtable}

\usepackage[serbian,english]{babel}

\usepackage{breakcites}
\usepackage{microtype}

\usepackage{soul}
\usepackage{wrapfig}

\usepackage{xcolor,colortbl}
\usepackage{pifont}
\newcommand{\cmark}{\ding{51}}%
\newcommand{\xmark}{\ding{55}}%

\usepackage{bm} 

\usepackage{centernot} 
\usepackage[shortlabels]{enumitem} 
\usepackage{amssymb}  

\usepackage{enumitem}

\usepackage{amsthm}  

\usepackage{mathtools} 
\usepackage{hyperref}

\newcommand{\E}{\mathbb{E}}

\usepackage{tikz} 
\usetikzlibrary{positioning}
\usepackage{tikz-dependency}
\usetikzlibrary{arrows}

\usepackage{hyperref}
\hypersetup{
	colorlinks,
	linkcolor={blue!70!black},
	citecolor={blue!70!black},
	urlcolor=black,
}

\title{On the Transfer of Inductive Bias\\ from Simulation to the Real World:\\ a New Disentanglement Dataset}

\author{%
	Muhammad Waleed Gondal$^{1  \ast}$\footnotemark[2]  
	 \And Manuel W\"uthrich$^{1}$\thanks{These authors contributed equally.}
	 \And {\DJ}or{\dj}e Miladinovi{\'c}$^{2}$
	 \And Francesco Locatello$^{12}$
	 \And Martin Breidt$^{3}$
	 \And Valentin Volchkov$^1$
	 \And Joel Akpo$^1$
	 \And Olivier Bachem$^4$
	 \And Bernhard Sch\"olkopf$^{1}$
	 \And Stefan Bauer$^1$\thanks{Correspondence to: \texttt{waleed.gondal@tue.mpg.de, stefan.bauer@inf.ethz.ch}}\\
	 \AND \\
	$^1$Max Planck Institute for Intelligent Systems\\
	$^2$Department of Computer Science ETH Zurich\\
	$^3$Max Planck Institute for Biological Cybernetics\\
	$^4$Google Research, Brain Team\\
}

\begin{document}

\maketitle

\begin{abstract}
Learning meaningful and compact representations with disentangled semantic aspects is considered to be of key importance in representation learning. Since real-world data is notoriously costly to collect, many recent state-of-the-art disentanglement models have heavily relied on synthetic toy data-sets. In this paper, we propose a novel data-set which consists of over one million images of physical 3D objects with seven factors of variation, such as object color, shape, size and position. In order to be able to control all the factors of variation precisely, we built an experimental platform where the objects are being moved by a robotic arm. In addition, we provide two more datasets which consist of simulations of the experimental setup. These datasets provide for the first time the possibility to systematically investigate how well different disentanglement methods perform on real data in comparison to simulation, and how simulated data can be leveraged to build better representations of the real world. We provide a first experimental study of these questions and our results indicate that learned models transfer poorly, but that model and hyperparameter selection is an effective means of transferring information to the real world.

\end{abstract}

\section{Introduction}
 In representation learning it is commonly assumed that a high-dimensional observation $\mathbf{X}$ is generated from  low-dimensional factors of variation  $\mathbf{G}$. The goal is usually to revert this process by searching for a latent embedding $\mathbf{Z}$ which replicates the underlying generative factors $\mathbf{G}$, e.g. shape, size or color. 
 Learning well-\emph{disentangled} representations of complex sensory data has been identified as one of the key challenges in the quest for artificial intelligence (AI) \cite{bengio2013representation, peters2017elements, lecun2015deep,bengio2007scaling,schmidhuber1992learning,lake2017building,vanSteenkiste2019abstract}, since they should contain all the information present in the observations in a compact and interpretable structure~\cite{bengio2013representation,kulkarni2015deep,chen2016infogan} while being independent from the task at hand~\cite{goodfellow2009measuring,lenc2015understanding}. 
 
Disentangled representations may be useful for (semi-)supervised learning of downstream tasks, transfer and few-shot learning~\cite{bengio2013representation,scholkopf2012causal, locatello2019disentangling}. Further, such representations allow to filter out nuisance factors~\cite{kumar2017variational}, to perform interventions and to answer counterfactual questions~\cite{pearl2009causality,SpiGlySch93,peters2017elements}. First applications of algorithms for learning disentangled representations have been applied to visual concept learning, sequence modeling, curiosity-based exploration or even domain adaptation in reinforcement learning  \cite{steenbrugge2018improving,laversanne2018curiosity,nair2018visual,higgins2017darla,higgins2018scan, lesort2018state,vanSteenkiste2019abstract}. The research community is in general agreement on the importance of this paradigm and much progress has been made in the past years, particularly on the algorithmic level~\cite[e.g.][]{higgins2017beta,kim2018disentangling}, fundamental understanding~\cite[e.g.][]{higgins2018towards,suter2018interventional} and experimental evaluation~\cite{locatello2018challenging}. However, research has thus far focused on synthetic toy datasets. 
 
The main motivation for using synthetic datasets is that they are cheap, easy to generate and the independent generative factors can be easily controlled. However, real-world recordings exhibit \emph{imperfections} such as chromatic aberrations in cameras and complex surface properties of objects (e.g. reflectance, radiance and irradiance), making transfer learning from synthetic to real data a nontrivial task. Despite the growing importance of the field and the potential societal impact in the medical domain or fair decision making \cite[e.g.][]{chartsias2018factorised,creager2019,locatello2019fairness}, the performance of state-of-the-art disentanglement learning on real-world data is unknown.
\begin{figure}[!h]
  \centering
  \begin{subfigure}[b]{.16\linewidth}
    \includegraphics[width=\textwidth]{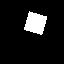}
  \end{subfigure}
  \begin{subfigure}[b]{.16\linewidth}
    \includegraphics[width=\textwidth]{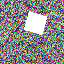}
  \end{subfigure}
  \begin{subfigure}[b]{.16\linewidth}
    \includegraphics[width=\textwidth]{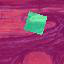}
  \end{subfigure}
  \begin{subfigure}[b]{.16\linewidth}
    \includegraphics[width=\textwidth]{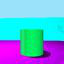}
  \end{subfigure}
  \begin{subfigure}[b]{.16\linewidth}
    \includegraphics[width=\textwidth]{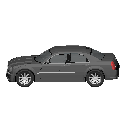}
  \end{subfigure}
  \begin{subfigure}[b]{.16\linewidth}
    \includegraphics[width=\textwidth]{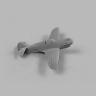}
  \end{subfigure}
    \begin{subfigure}[b]{.325\linewidth}
    \includegraphics[width=\textwidth]{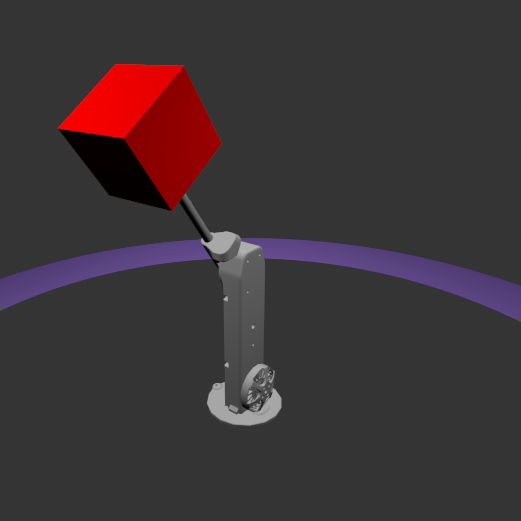}
  \end{subfigure}
  \begin{subfigure}[b]{.325\linewidth}
    \includegraphics[width=\textwidth]{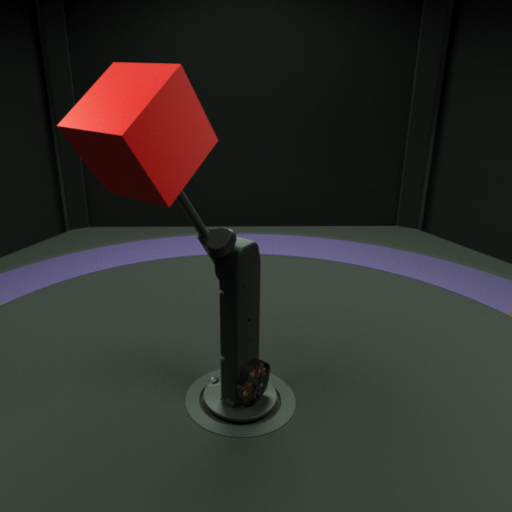}
  \end{subfigure}
  \begin{subfigure}[b]{.325\linewidth}
    \includegraphics[width=\textwidth]{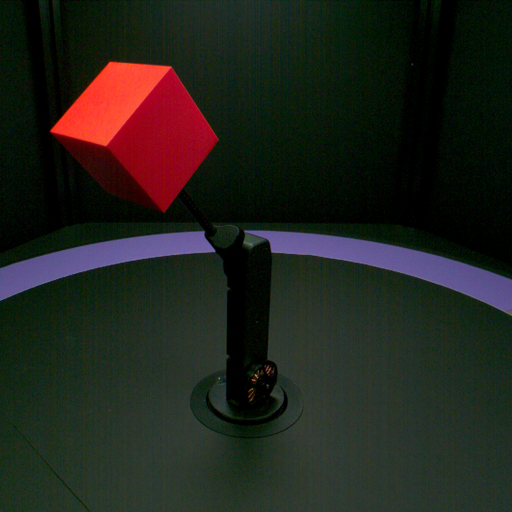}
  \end{subfigure}
 
  \caption{\textbf{Datasets:} In the top row samples from previously published datasets are shown from left to right: dSprites, Noisy-dSprites, Scream-dSprites, 3dshapes, Cars3D and SmallNORB. In the second row (again left to right) we provide simple simulated data, highly realistically simulated data and real-world data examples of the newly collected dataset.}
  \label{fig:datasets}
\end{figure}

To bridge the gap between simulation and the physical world, we built a recording platform which allows to investigate the following research questions: (i) How well do unsupervised state-of-the-art algorithms transfer from rendered images to physical recordings? (ii) How much does this transfer depend on the quality of the simulation? (iii) Can we learn representations on low dimensional recordings and transfer them from the current state-of-the-art of $64 \times 64$ images to high quality images? (iv) How much supervision is necessary to encode the necessary inductive biases? (v) Are the confounding and distortions of real-world recordings beneficial for learning disentangled representations? (vi) Can we disentangle causal mechanisms  \citep{pearl2009causality,lake2015human, lake2017building, peters2017elements} in the data generating process? (vii) Are disentangled representations useful for solving the real-world downstream tasks?

While answering all of the above questions is beyond the scope of this paper, our key contributions can be summarized as follows: 

  \setlist{nolistsep}
\begin{itemize}
\itemsep0em 
    \item We introduce the first \emph{real-world 3D data set} recorded in a controlled environment, defined by \emph{7 factors of variation}: object color, object shape, object size, camera height, background color and two degrees of freedom of motion of a robotic arm. The dataset is made publicly available\footnote{\url{https://github.com/rr-learning/disentanglement_dataset}}.
    \item We provide synthetic images produced by computer graphics with two levels of realism. Since the robot arm and the objects are printed from a 3D template, we can ensure a close similarity between the realistic renderings and the real-world recordings.
    \item The collected datatset of physical 3D objects consists of over one million images, and each of the two simulated datasets contains the same number of images as well.
    \item We investigate the role of inductive bias and the transfer of different hyper-parameter settings between the different simulations and the real-world and the requirements on the quality of the simulation for a succesful transfer.
  \end{itemize}

\section{Background and Related Work}
\label{sec:related_work}

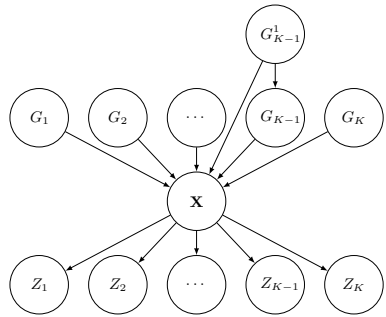
\begin{wrapfigure}{r}{5cm}
  \resizebox{0.8\totalheight}{!}{
          \begin{tikzpicture}[
            node distance=0.5 cm and 0.4 cm,
            mynode/.style={draw,circle, align=center, inner sep=0pt, text width=6mm, minimum size = 1.2cm},
            ]
            \node[mynode](g1){\small $G_1$};
            \node[mynode, right =of g1] (g2){\small $G_2$};
            \node[mynode, right =of g2](dots){\small $\cdots$};
            \node[mynode, right =of dots](gkm1){\small $G_{K-1}$};
            \node[mynode, right =of gkm1](gk){\small $G_K$};
             \node[mynode, above =of gkm1](gk1){\small $G_{K-1}^1$};
            \node[mynode, below  =of dots] (x) {\small $\mathbf{X}$};
            \node[mynode, below =of x](zdots){\small $\cdots$};
            \node[mynode, left =of zdots] (z2){\small $Z_2$};
            \node[mynode, left = of z2](z1){\small $Z_1$};
            \node[mynode, right =of zdots](zkm1){\small $Z_{K-1}$}; 
            \node[mynode, right =of zkm1](zk){\small $Z_{K}$};
            \path
                      (gk1) edge[-latex] (x)
                       (gk1) edge[-latex] (gkm1)
            (g1) edge[-latex] (x)
            (g2) edge[-latex] (x)
            (dots) edge[-latex] (x)
            (gkm1) edge[-latex] (x)
            (gk) edge[-latex] (x)
            (x) edge[-latex] (z1)
            (x) edge[-latex] (z2)
            (x) edge[-latex] (zdots)
            (x) edge[-latex] (zkm1)
            (x) edge[-latex] (zk)
            ;
            \end{tikzpicture}
            }
      \caption{Graphical Model, where $\mathbf{G} = (G_1, G_2, \dots, G_K)$ are the generative factors (color, shape, size, ...) and $\mathbf{X}$ the recorded images. The aim of disentangled representation learning is to learn variables $Z_i$ that capture the independent mechanisms $G_i$.}
      \label{fig:causal-structure}
\end{wrapfigure}

We assume a set of observations of a (potentially high dimensional) random variable $\bm{X}$ which is generated by $K$ unobserved causes of variation (generative factors) $\bm{G} = [G_1,\dots,G_K]$ (i.e., $\bm{G} \rightarrow \bm{X}$) that do not cause each other.  These latent factors represent \textit{elementary ingredients} to the causal mechanism generating $\bm{X}$ \cite{pearl2009causality, peters2017elements}. The elementary ingredients $G_i, i=1,\dots,K$ of the causal process work on their own and are changeable without affecting others, reflecting the independent mechanisms assumption \cite{scholkopf2012causal}. However, for some of the factors a hierarchical structure may exist for which this may only hold true when seeing the hierachical structure as a whole as one component. The graphical model corresponding to this framework and adapted from \cite{suter2018interventional} is depicted in figure \ref{fig:causal-structure}. The hirachical structure of the factors $G_{K-1}^1$  and  $G_{K-1}$ might represent one compositional process e.g. connected joints of a robot arm.  The most commonly accepted understanding of \emph{disentanglement} \cite{bengio2013representation} is that each learned feature in $\bm{Z}$ should capture one factor of variation in $\bm{G}$.

\if{0}
\begin{figure}[htp!]
    \begin{subfigure}{.49\textwidth}
          \centering
          \includegraphics[width=.98\linewidth]{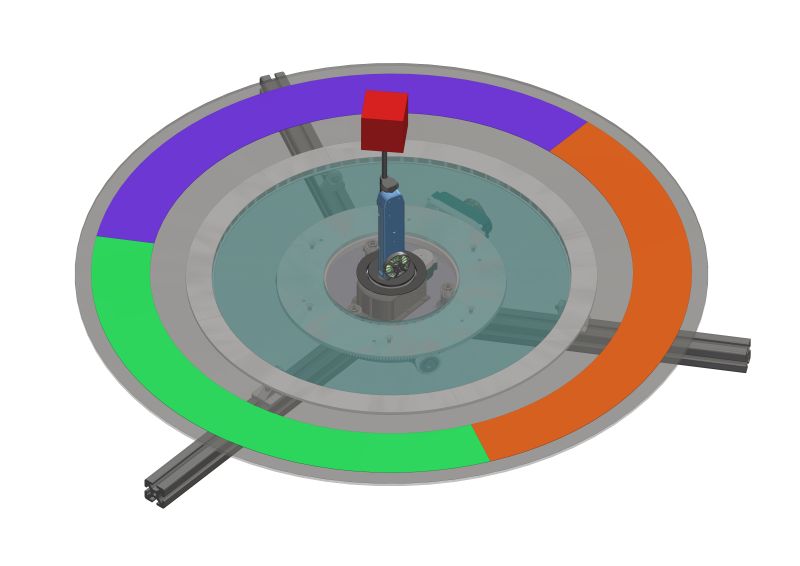}
          \label{fig:sfig2}
    \end{subfigure}
    \begin{subfigure}{.49\textwidth}
    \resizebox{0.9\totalheight}{!}{
          \begin{tikzpicture}[
            node distance=0.5 cm and 0.4 cm,
            mynode/.style={draw,circle, align=center, inner sep=0pt, text width=6mm, minimum size = 1.2cm},
            ]
            \node[mynode](g1){\small $G_1$};
            \node[mynode, right =of g1] (g2){\small $G_2$};
            \node[mynode, right =of g2](dots){\small $\cdots$};
            \node[mynode, right =of dots](gkm1){\small $G_{K-1}$};
            \node[mynode, right =of gkm1](gk){\small $G_K$};
             \node[mynode, above =of gkm1](gk1){\small $G_{K-1}^1$};
            \node[mynode, below  =of dots] (x) {\small $\mathbf{X}$};
            \node[mynode, below =of x](zdots){\small $\cdots$};
            \node[mynode, left =of zdots] (z2){\small $Z_2$};
            \node[mynode, left = of z2](z1){\small $Z_1$};
            \node[mynode, right =of zdots](zkm1){\small $Z_{K^\prime-1}$};
            \node[mynode, right =of zkm1](zk){\small $Z_{K^\prime}$};
            \path
                      (gk1) edge[-latex] (x)
                       (gk1) edge[-latex] (gkm1)
            (g1) edge[-latex] (x)
            (g2) edge[-latex] (x)
            (dots) edge[-latex] (x)
            (gkm1) edge[-latex] (x)
            (gk) edge[-latex] (x)
            (x) edge[-latex] (z1)
            (x) edge[-latex] (z2)
            (x) edge[-latex] (zdots)
            (x) edge[-latex] (zkm1)
            (x) edge[-latex] (zk)
            ;
            \end{tikzpicture}
            }
    \end{subfigure}
    \caption{\textbf{Disentangled Causal Mechanisms}  On the left the robotic arm with two connected joints is shown, while on the right a graphical model corresponding to the causal framework \cite{suter2018interventional, bengio2019meta} for learning disentangled representations is shown. $\mathbf{G} = (G_1, G_2, \dots, G_K)$ are the generative factors (shape, size, background color, ...) and $\mathbf{X}$ the recorded images. While simulated recordings are not confounded by $C$, it is even possible that multiple confounder exists for the real-world recordings. The factors $G_{K-1}$ and $G_{K-1}^1$ would correspond to the hierarchical structure of the first and the second joint. The aim of disentangled representation learning is to learn variables $Z_i$ that capture the independent mechanisms $G_i$. }
    \label{fig:causal-structure}
    \vspace{-2cm}
\end{figure}
\fi

Current state-of-the-art disentanglement approaches use the framework of variational auto-encoders (VAEs) \cite{kingma2013auto}. The (high-dimensional) observations $\bm{x}$ are modelled as being generated from some latent features $\bm{z}$ with chosen prior $p(\bm{z})$ according to the probabilistic model $p_\theta(\bm{x}|\bm{z})p(\bm{z})$. The generative model $p_\theta(\bm{x}|\bm{z})$ as well as the proxy posterior $q_\phi(\bm{z}|\bm{x})$ can be represented by neural networks, which are optimized by maximizing the variational lower bound (ELBO) of $\log p(\bm{x}_1,\dots,\bm{x}_N)$

$\mathcal{L}_{VAE} = \sum_{i=1}^N  \E_{q_\phi(\bm{z}|\bm{x}^{(i)})} [ \log p_\theta (\bm{x}^{(i)} | \bm{z}) ] 
- D_{KL}( q_\phi(\bm{z}|\bm{x}^{(i)}) \| p (\bm{z}) )$

 Since the above objective does not enforce any structure on the latent space, except for some similarity to the typically chosen isotropic Gaussian prior $p (\bm{z})$, various proposals for more structure imposing regularization have been made. Using some sort of supervision \cite[e.g.][]{narayanaswamy2017learning, bouchacourt2017multi, liu2017detach, mathieu2016disentangling, cheung2014discovering} or proposing completely unsupervised \cite[e.g.][]{higgins2016beta, kim2018disentangling, chen2018isolating, kumar2017variational, esmaeili2018hierarchical} learning approaches. \cite{higgins2016beta} proposed the $\beta$-VAE penalizing the Kullback-Leibler divergence (KL) term in the VAE objective  more strongly, which encourages similarity to the factorized prior distribution. Others used techniques to encourage statistical independence between the different components in $\bm{Z}$, e.g., FactorVAE \cite{kim2018disentangling} or $\beta$-TCVAE \cite{chen2018isolating}, while DIP-VAE proposed to encourage factorization of the inferred prior $q_\phi(\bm{z}) = \int q_\phi (\bm{z} | \bm{x}) p(\bm{x}) \, d\bm{x}$. For other related work we refer to the detailed descriptions in the recent empirical study \cite{locatello2018challenging}.

\subsection{Established Datasets for the Unsupervised Learning of Disentangled Representations}
\label{sec:datasets}

Real-world data is costly to generate and groundtruth is often not available since significant confounding may exist. To bypass this limitation, many recent state-of-the-art disentanglement models \cite{watters2019spatial, kim2018disentangling,chen2018isolating,higgins2017beta, chen2016infogan} have heavily relied on synthetic toy datasets, trying to solve a simplified version of the problem in the hope that the conclusions drawn might likewise be valid for real-world settings. A quantitative summary of the most widely used datasets for learning disentangled representations is provided in table \ref{tab:datasets}. 


\begin{table}[htb!]
	\centering
	\begin{tabular}{lccccc}
	\toprule
	Dataset &   \shortstack{Factors of \\ Variation}  & Resolution & 	\# of Images  & 3D & Real-World\\
\midrule
dSprites 	 & 5   & $64 \times 64$  & 737,280  & \textcolor{red}{\xmark}   & \textcolor{red}{\xmark} \\
Noisy dspirtes & 5   & $64 \times 64$ & 737,280 & \textcolor{red}{\xmark} & \textcolor{red}{\xmark} \\
Scream dSprites	 & 5   & $64 \times 64$ & 737,280 & \textcolor{red}{\xmark}  & \textcolor{red}{\xmark}\\
SmallNORB &  5   & $128 \times 128$ & 48,600 & \textcolor{green}{\cmark} & \textcolor{red}{\xmark} \\
Cars3D &  3    & $64 \times 64$  & 17,568 & \textcolor{green}{\cmark}  & \textcolor{red}{\xmark} \\
3dshapes	 & 6   & $64 \times 64$ & 480,000 & \textcolor{green}{\cmark}   & \textcolor{red}{\xmark}\\
\emph{MPI3D-toy}	& 7   & $64 \times 64$ & 1,036,800 & \textcolor{green}{\cmark} & \textcolor{red}{\xmark}  \\
\emph{MPI3D-realistic}	 & 7  & $256 \times 256$ & 1,036,800 & \textcolor{green}{\cmark} & \textcolor{red}{\xmark} \\
\emph{MPI3D-real}	 & 7   & $512 \times 512$ & 1,036,800 & \textcolor{green}{\cmark}  & \textcolor{green}{\cmark}  \\
	\bottomrule
	\end{tabular}
\vspace{0.3cm}
    \caption{Summary of the properties of different datasets. The newly contributed datasets are \emph{emphasized}. }
\label{tab:datasets}
\end{table}
\paragraph{Dataset Descriptions:} For quantitative analysis, \emph{dSprites}\footnote{\url{https://github.com/deepmind/dsprites-dataset}} is the most commonly used dataset. This synthetic dataset \cite{higgins2017beta} contains binary 2D images of hearts, ellipses and squares in low resolution. In \emph{Color-dSprites} the shapes are colored with a random color, \emph{Noisy-dSprites} considers white-colored shapes on a noisy background and in \emph{Scream-dSprites} the background is replaced with a random patch in a random color shade extracted from the famous The Scream painting \cite{munch_1893}. The dSprites shape is embedded into the image by inverting the color of its pixel. The \emph{SmallNORB}\footnote{\url{https://cs.nyu.edu/~ylclab/data/norb-v1.0-small/}} dataset contains images of 50 toys belonging to 5 generic categories: four-legged animals, human figures, airplanes, trucks, and cars. The objects were imaged by two cameras under 6 lighting conditions, 9 elevations (30 to 70 degrees every 5 degrees), and 18 azimuths (0 to 340 every 20 degrees) \cite{lecun2004learning}. For \emph{Cars3D}\footnote{\url{http://www.scottreed.info/files/nips2015-analogy-data.tar.gz}},199 CAD models from \cite{fidler20123d} were used to generate 64x64 color renderings from 24 rotation angles each offset by 15 degrees \cite{reed2015deep}. Recently, \emph{3dshapes} was made publicly available\footnote{\url{https://github.com/deepmind/3dshapes-dataset/}}, a dataset of 3D shapes procedurally generated from 6 ground truth independent latent factors. These factors are floor colour, wall colour, object colour, scale, shape and orientation \cite{kim2018disentangling}.

\section{Bridging the Gap Between Simulation and the Real World: A Novel Dataset}
\label{sec:novel_data}
While other real-world recordings, e.g.  \emph{CelebA} \cite{liu2015deep}, exist, they offer only qualitative evaluations. However, a more controlled dataset is needed to  quantitatively investigate the effects of inductive biases, sample complexity and the interplay of simulations and the  real-world. 

\subsection{Controlled Recording Setup}\label{sec:platform}
\begin{figure}[htp!]
    \centering
      \begin{subfigure}[b]{.32\linewidth}
    \includegraphics[width=\textwidth]{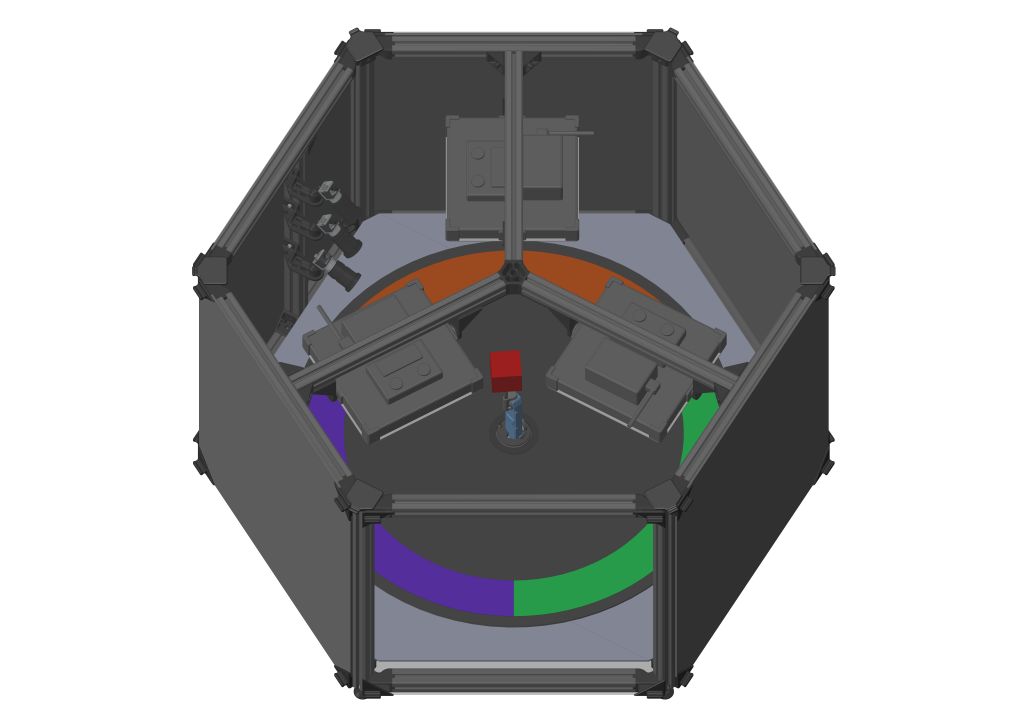}
      \end{subfigure}
          \begin{subfigure}[b]{.32\linewidth}
    \includegraphics[width=\textwidth]{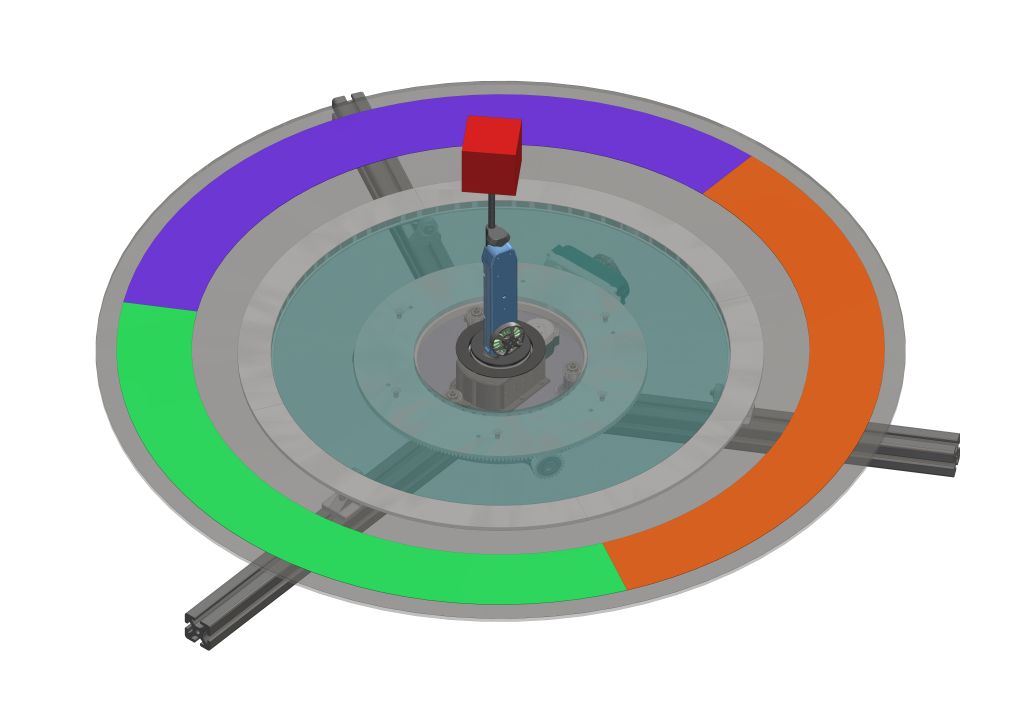}
  \end{subfigure}
      \begin{subfigure}[b]{.32\linewidth}
    \includegraphics[width=\textwidth]{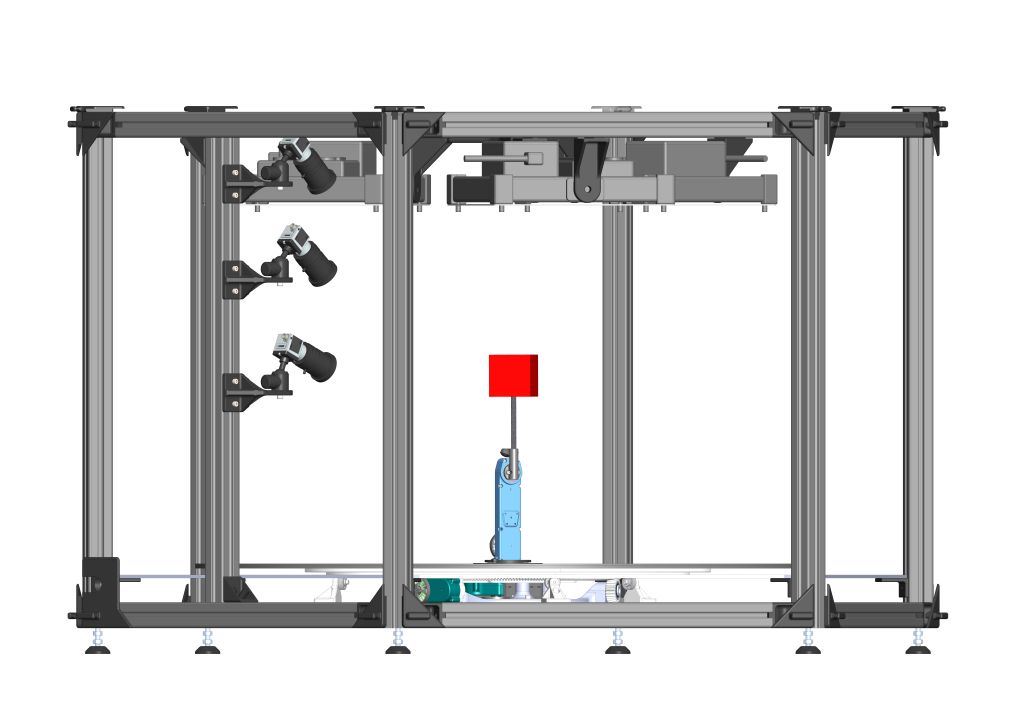}
      \end{subfigure}

    \caption{Renderings of the developed robotic platform: On the left a view from a 30$^\circ$ angle from the top (note that one panel in front and the top panels have been removed such that the interior of the platform is visible. During recordings, the platform is entirely closed. Middle: the robotic arm carrying a red cube (the entire cage is hidden). Right: frontal view without the black shielding (note the three cameras at different heights). }
        \label{fig:setup}
\end{figure}%
In order to record a controlled dataset of physical 3D objects, we built the mechanical platform illustrated in figure~\ref{fig:setup}. It consists of three cameras mounted at different heights, a robotic manipulator carrying a 3D printed object (which can be swapped) in the center of the platform and a rotating table at the bottom. The platform is shielded with black sheets from all sides to avoid any intrusion of external factors (e.g. light)  and the whole environment is relatively uniformly illuminated by three light sources installed within the platform.

\subsubsection{Factors of Variation}

 The generative factors of variation $\bm{G}$ mentioned in section \ref{sec:related_work} are listed in the following for our recording setup.

\paragraph{Object Color:}

\begin{figure}[!htp]
	\centering
	\begin{subfigure}[b]{.16\linewidth}
		\includegraphics[width=\textwidth]{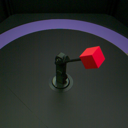}
	\end{subfigure}
	\begin{subfigure}[b]{.16\linewidth}
		\includegraphics[width=\textwidth]{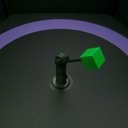}
	\end{subfigure}
	\begin{subfigure}[b]{.16\linewidth}
		\includegraphics[width=\textwidth]{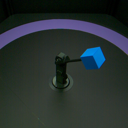}
	\end{subfigure}
	\begin{subfigure}[b]{.16\linewidth}
		\includegraphics[width=\textwidth]{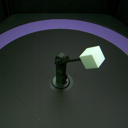}
	\end{subfigure}
	\begin{subfigure}[b]{.16\linewidth}
		\includegraphics[width=\textwidth]{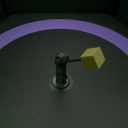}
	\end{subfigure}
	\begin{subfigure}[b]{.16\linewidth}
		\includegraphics[width=\textwidth]{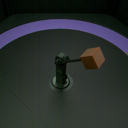}
	\end{subfigure}
	
	\caption{We show all the object colors while maintaining the other factors constant.}
	\label{fig-object-colors}
\end{figure}

All objects have one of six different colors: red (255, 0, 0), green (0, 255, 0), blue (0, 0, 255), white (255, 255, 255), olive (210,210,80) and brown (153,76,0) (see figure~\ref{fig-object-colors}). 

\begin{figure}[!htb]
  \centering
  \begin{subfigure}[b]{.16\linewidth}
    \includegraphics[width=\textwidth]{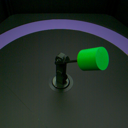}
  \end{subfigure}
  \begin{subfigure}[b]{.16\linewidth}
    \includegraphics[width=\textwidth]{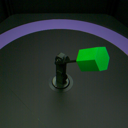}
  \end{subfigure}
  \begin{subfigure}[b]{.16\linewidth}
    \includegraphics[width=\textwidth]{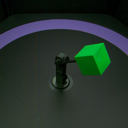}
  \end{subfigure}
  \begin{subfigure}[b]{.16\linewidth}
    \includegraphics[width=\textwidth]{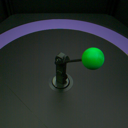}
  \end{subfigure}
  \begin{subfigure}[b]{.16\linewidth}
    \includegraphics[width=\textwidth]{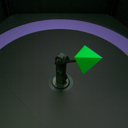}
  \end{subfigure}
  \begin{subfigure}[b]{.16\linewidth}
    \includegraphics[width=\textwidth]{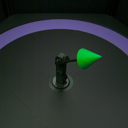}
  \end{subfigure}
  
  \caption{We show all object shapes while maintaining all other factors constant.}
  \label{fig-object-shapes}
\end{figure}

  \begin{wrapfigure}{l}{5cm}
     \begin{tabular}{@{}cc@{}}
   \includegraphics[width=0.16\textwidth]{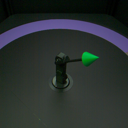}&
   \includegraphics[width=0.16\textwidth]{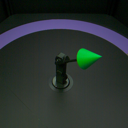}\\
     \end{tabular}
       \caption{We show the two object sizes while maintaining all other factors constant.}
       \label{fig:object-sizes}
 \end{wrapfigure}

\paragraph{Object Shape:}
There are objects of four different shapes in the dataset: a cylinder, a hexagonal prism, a cube, a sphere, a pyramid with square base and a cone. All objects exhibit rotational symmetries about some axes, however the kinematics of the robot are such that these axes never align with the degrees of freedom of the robot. This is important because it ensures that the robot degrees of freedom are observable given the images.
 
 \paragraph{Object Size:}
There are objects of two different sizes in the dataset, categorized as large (roughly 65mm in diameter) and small (roughly 45 mm in diameter).

\paragraph{Camera Height:}
\begin{figure}[tb]
	\centering
	\begin{subfigure}[b]{.16\linewidth}
		\includegraphics[width=\textwidth]{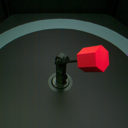}
	\end{subfigure}
	\begin{subfigure}[b]{.16\linewidth}
		\includegraphics[width=\textwidth]{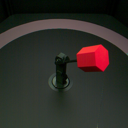}
	\end{subfigure}
	\begin{subfigure}[b]{.16\linewidth}
		\includegraphics[width=\textwidth]{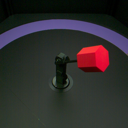}
	\end{subfigure}
	\begin{subfigure}[b]{.16\linewidth}
		\includegraphics[width=\textwidth]{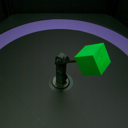}
	\end{subfigure}
	\begin{subfigure}[b]{.16\linewidth}
		\includegraphics[width=\textwidth]{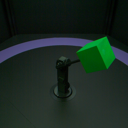}
	\end{subfigure}
	\begin{subfigure}[b]{.16\linewidth}
		\includegraphics[width=\textwidth]{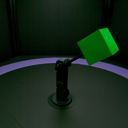}
	\end{subfigure}
	
	\caption{Three images on the left: we vary the backround color. Three images on the right: we vary the camera height.}
	\label{fig-bg-strip-and-heights}
\end{figure}
The dataset is recorded with three cameras mounted at three different heights (see figure~\ref{fig-bg-strip-and-heights} on the right), which represents another factor of variation in the images.

\paragraph{Background Color:}
The rotation table (see figure~\ref{fig-bg-strip-and-heights}) allows us to change background color. Note that for all images in the dataset we orient the table in such a way that only one background color is visible at a time. The colors are: sea green, salmon and purple.

\begin{figure}[ht!]
  \centering
  \begin{subfigure}[b]{.16\linewidth}
    \includegraphics[width=\textwidth]{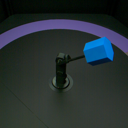}
  \end{subfigure}
  \begin{subfigure}[b]{.16\linewidth}
    \includegraphics[width=\textwidth]{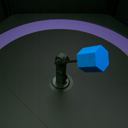}
  \end{subfigure}
  \begin{subfigure}[b]{.16\linewidth}
    \includegraphics[width=\textwidth]{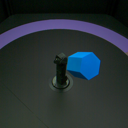}
  \end{subfigure}
  \begin{subfigure}[b]{.16\linewidth}
    \includegraphics[width=\textwidth]{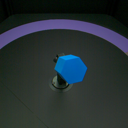}
  \end{subfigure}
  \begin{subfigure}[b]{.16\linewidth}
    \includegraphics[width=\textwidth]{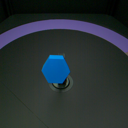}
  \end{subfigure}
  \begin{subfigure}[b]{.16\linewidth}
    \includegraphics[width=\textwidth]{figures/factors_real_world/first_dof/firstDOF_B_4.png}
  \end{subfigure}
  
  \caption{Motion along first DOF while maintaining the other factors constant. Note that in total we record 40 steps, here we only show 6 due to space constraints.}
  \label{fig-first-dof}
\end{figure}

\begin{figure}[ht!]
  \centering
  \begin{subfigure}[b]{.16\linewidth}
    \includegraphics[width=\textwidth]{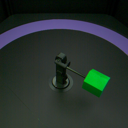}
  \end{subfigure}
  \begin{subfigure}[b]{.16\linewidth}
    \includegraphics[width=\textwidth]{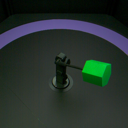}
  \end{subfigure}
  \begin{subfigure}[b]{.16\linewidth}
    \includegraphics[width=\textwidth]{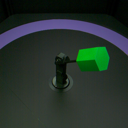}
  \end{subfigure}
  \begin{subfigure}[b]{.16\linewidth}
    \includegraphics[width=\textwidth]{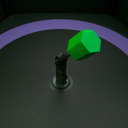}
  \end{subfigure}
  \begin{subfigure}[b]{.16\linewidth}
    \includegraphics[width=\textwidth]{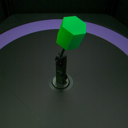}
  \end{subfigure}
  \begin{subfigure}[b]{.16\linewidth}
    \includegraphics[width=\textwidth]{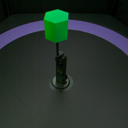}
  \end{subfigure}
  \if{0}
  \begin{subfigure}[b]{.16\linewidth}
    \includegraphics[width=\textwidth]{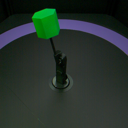}
  \end{subfigure}
  \begin{subfigure}[b]{.16\linewidth}
    \includegraphics[width=\textwidth]{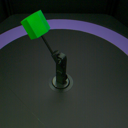}
  \end{subfigure}
  \begin{subfigure}[b]{.16\linewidth}
    \includegraphics[width=\textwidth]{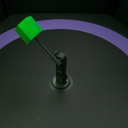}
  \end{subfigure}
  \begin{subfigure}[b]{.16\linewidth}
    \includegraphics[width=\textwidth]{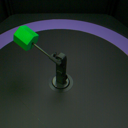}
  \end{subfigure}
  \begin{subfigure}[b]{.16\linewidth}
    \includegraphics[width=\textwidth]{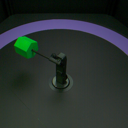}
  \end{subfigure}
\fi
  \caption{Motion along second DOF while maintaining the other factors constant. Note that in total we record 40 steps, here we only show 6 due to space constraints.}
  \label{fig-second-dof}
\end{figure}

\paragraph{Degrees of Freedom of the Robotic Arm:}
Each object is mounted on the tip of the manipulator shown in figure~\ref{fig:setup}. This manipulator has two degrees of freedom, a rotation about a vertical axis at the base and a second rotation about a horizontal axis.
We move each joint in a range of 180$^{\circ}$ in 40 equal steps (see figure~\ref{fig-first-dof} and figure~\ref{fig-second-dof}). Note that these two factors of variation are independent, just like all other factors (i.e. we record all possible combinations between the two).

\subsection{Simulated Data}

In addition to the real-world dataset we recorded two simulated datasets of the same setup, hence all factors of variation are identical across the three datasets. One of the simulated datasets is designed to be as realistic as possible and the synthetic images are visually practically indistinguishable from real images (see figure~\ref{fig:datasets} middle). For the second simulated dataset we used a  deliberately simplified model (see figure~\ref{fig:datasets} left), which allows to investigate transfer from simplified models to real data.

The synthetic data was generated using Autodesk 3ds Max(2018). Most parts of the scene were imported from SolidWorks CAD files that were originally used to construct the experimental stage including the manipulator and 3D printing of the test objects. The surface shaders are based on Autodesk Physical material with hand-tuned shading parameters, based on full resolution reference images. The camera poses were initialized from the CAD data and then manually fine-tuned using reference images. The realistic synthetic images were obtained using the Autodesk Raytracer (ART) with three rectangular light sources, mimicking the LED panels. The simplified images were rendered with the Quicksilver hardware renderer.

\section{First Experimental Evaluations of (unsupervised) Disentanglement Methods on Real-World Data}
\label{sec:experimental_setup}

Some fields have been able to narrow the gap between simulation and reality \cite{zhang2019vr, bousmalis2016domain, james2018sim}, which has led to remarkable achievements (e.g. for in-hand manipulation \cite{andrychowicz2018learning}). In contrast, for disentanglement methods this gap has not been bridged yet, state-of-the-art algorithms seem to have difficulties to transfer learned representations even between toy datasets \cite{locatello2018challenging}. The proposed dataset will enable the community to systematically investigate how such transfer of information between simulations with different degrees of realism and real data can be achieved. In the following we present a first experimental study in this directions.

\subsection{Experimental Protocol} 
We apply all the disentanglement methods ($\beta$-VAE, FactorVAE, $\beta$-TCVAE, DIP-VAE-I, DIP-VAE-II, AnnealedVAE) which were used in a recent large-scale study \cite{locatello2018challenging} to our three datasets. Due to space constraints, the models are abbreviated with numbers one to five in the plots in the same order. We use (\texttt{disentanglement\textunderscore lib}) and we evaluate on the same scores as \cite{locatello2018challenging}. In all the experiments, we used images with resolution 64x64. This resolution is used in the recent large-scale evaluations and by state-of-the-art disentanglement learning algorithms \cite{locatello2018challenging}. Each of the six methods is trained on each of the three datasets with five different hyperparameter settings (see table~\ref{tab:models} in the appendix for details) and with three different random seeds, leading to a total of 270 models. Each model is trained for 300,000 iterations on Tesla V100 gpus. Details about the evaluation metrics can be found in appendix~\ref{sec:appendix_metrics}.

\subsection{Experimental Results}

 \begin{figure}[htp!] 
 	\centering	
 	\includegraphics[width=\textwidth, height=3.5cm]{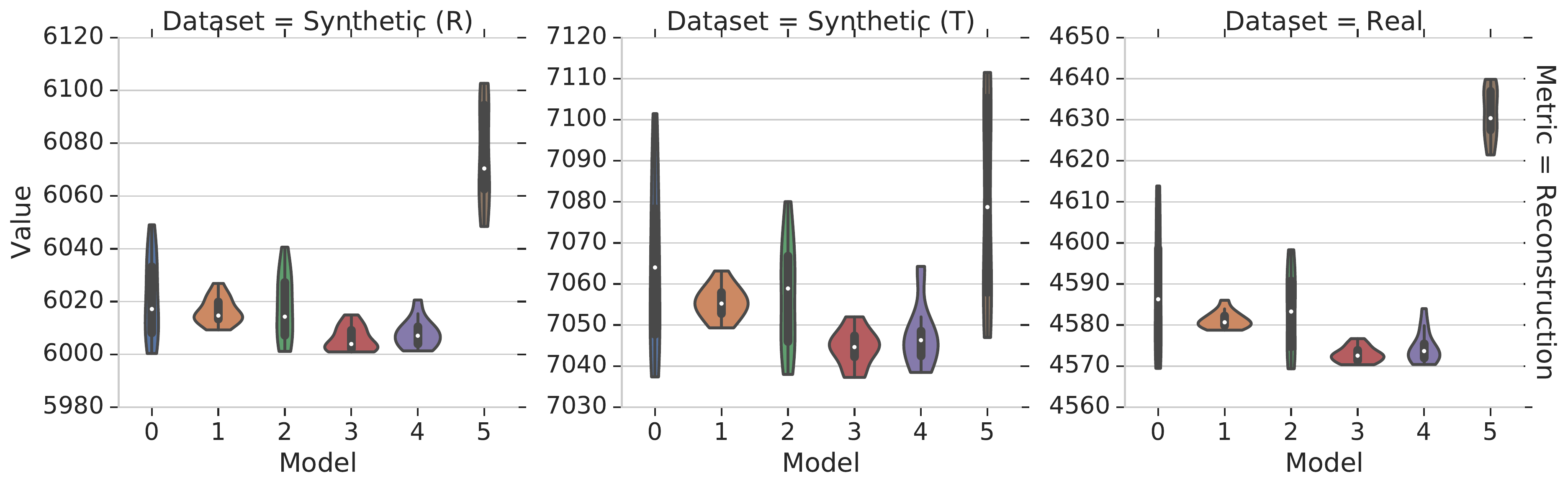}
 	\caption{Reconstruction scores of different methods (0=$\beta$-VAE, 1=FactorVAE, 2=$\beta$-TCVAE, 3=DIP-VAE-I, 4=DIP-VAE-II, 5=AnnealedVAE) on the realistic synthetic dataset, the toy synthetic dataset and the real dataset.}
 	\label{fig:reconstruction_complexity}
 \end{figure}
 
 \textbf{Reconstruction Across Datasets:} Figure~\ref{fig:reconstruction_complexity} shows that there is a difference in reconstruction score across datasets: The score is the lowest on real data, followed by the realistic simulated dataset (R) and the simple toy (T) images. This indicates that there is a significant difference in the distribution of the real data compared to the simulated data, and that it is harder to learn a representation of the real data than of the simulated data. However, the relative behaviour of different methods seems to be similar across all three datasets, which indicates that despite the differences, the simulated data may be useful for model selection.

\paragraph{Direct Transfer of Representations:} 
 \begin{figure}[htp!] 
 	\centering	
 	\includegraphics[width=\textwidth]{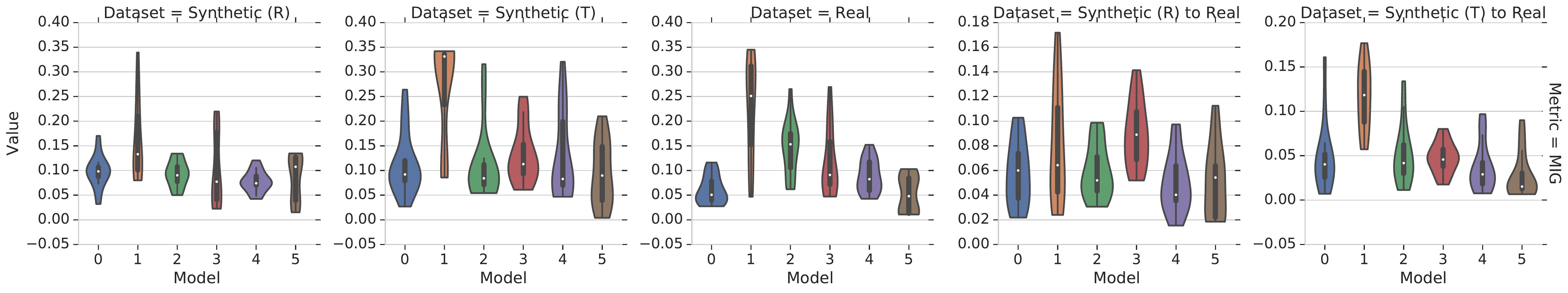} 
 	\caption{The Mutual Information Gap (MIG) scores attained by different methods for the following evaluations (from left to right): (a) trained and evaluated on synthetic realistic, (b) trained and evaluated on synthetic toy, (c) trained and evaluated on real, (d) trained on synthetic realistic and evaluated on real, (e) trained on synthetic toy and evaluated on real. The variance is due to different hyper-parameters and seeds. 
 	}
 	\label{fig:mig_violin}
 \end{figure}
In figure~\ref{fig:mig_violin} we show the Mutual Information Gap (MIG) scores attained by different methods for different evaluations. The same plots for different metrics look qualitatively similar (see figure~\ref{fig:appendix_violin} in the appendix). Given the high variance, it is difficult to make conclusive statements. However, it seems quite clear that all methods perform significantly better when they are trained and evaluated on the same dataset (three plots on the left). Direct transfer of learned representations from simulated to real data (two plots on the right) seems to work rather poorly.

\paragraph{Transfer of Hyperparameters:} 
\begin{wrapfigure}{l}{0.3\textwidth}
     \vskip 0.2in
 	\centering	
 	\includegraphics[width=0.3\textwidth]{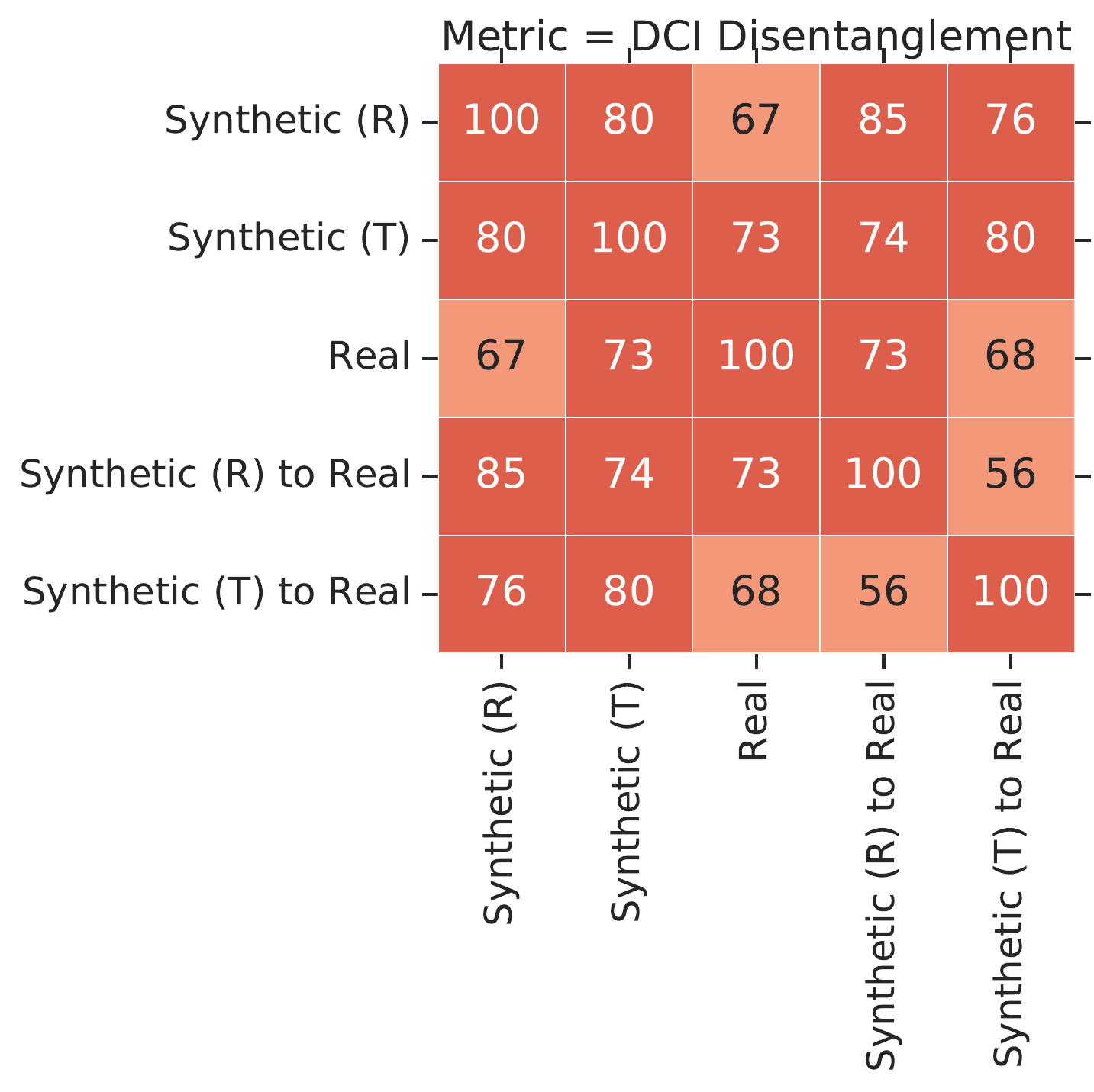}
 	\caption{Rank-correlation of the DCI disentanglement scores of different models (including hyperparameters) across different data sets.}
 	\label{fig:transfer_dci}
\end{wrapfigure}
We have seen that transferring representations directly from simulated to real data seems to work poorly. However, it may be possible to instead transfer information at a higher-level, such as the choice of the method and its hyperparameters as an inductive bias.

In order to quantitatively evaluate whether such a transfer is possible, we pick the model (including hyperparameters) which performs best in simulation (according to a metric chosen at random), and we compute the probability of outperforming (according to a metric and seed chosen randomly) a model which was chosen at random. If no transfer is possible, we would expect this probability to be $50\%$.

However, we find that model selection from realistic simulated renderings (R) outperforms random model selection $72\%$ of the time while transferring the model from the simpler synthetic images (T) to real-world data even beats random selection $78\%$ of the time.

This finding is confirmed by figure~\ref{fig:transfer_dci}, where we show the rank-correlation of the performance of models (including hyperparameters) trained on one dataset with the performance of these models trained on another dataset. The performance of a model trained on some dataset seems to be highly correlated with the performance of that model trained on any other dataset. In figure~\ref{fig:transfer_dci} we use the DCI disentanglement metric as a score, however, qualitatively similar results can be observed using most of the disentanglement metrics (see figure~\ref{fig:rank_across} in the appendix).

\textbf{Summary} These results indicate that the simulated and the real data distribution have some similarities, and that these similarities can be exploited through model and hyperparameter selection. Surprisingly, it seems that the transfer of models from the synthetic toy dataset may work even better than the transfer from the realistic synthetic dataset.

\section{Conclusions}
Despite the intended applications of disentangled representation learning algorithms to real data in fields such as robotics, healthcare and fair decision making \cite[e.g.][]{chartsias2018factorised,creager2019, higgins2017darla}, state-of-the-art approaches have only been systematically evaluated on synthetic toy datasets. Our work effectively complements related efforts \cite[e.g.][]{locatello2018challenging} to address current challenges of representation learning, offering the possibility of investigating the role of inductive biases,  sample complexity, transfer learning and the use of labels 
using real-world images.

A key aspect of our datasets is that we provide rendered images of increasing complexity for the \emph{same} setup used to capture the real-world recordings. 
The different recordings offer the possibility of investigating the question if disentangled representations can be transferred from  simulation to the real world and how the transferability depends on the degree of \emph{realism} of the simulation. 
Beyond the evaluation of representation learning algorithms, the proposed dataset can likewise be used for other tasks such as 3D reconstruction and scene rendering \cite{eslami2018neural} or learning compositional visual concepts \cite{higgins2017scan}. Furthermore, we are planning to use the novel experimental setup for recording objects with more complicated shapes and textures under more difficult conditions, such as dependence among different factors.

\subsubsection*{Acknowledgments}
This research was partially supported by the Max Planck ETH Center for Learning Systems and Google Cloud. We thank Alexander Neitz and Arash Mehrjou for useful discussions. We would also like to thank Felix Grimminger, Ludovic Righetti, Stefan Schaal, Julian Viereck and Felix Widmaier whose work served as a starting point for the development of the robotic platform in the present paper.

\clearpage

\bibliographystyle{plain}
\bibliography{references}

\clearpage

\appendix
\section{Platform}
\label{sec:appendix_platform}

\begin{figure}[htp!]
    \centering
    
      \begin{subfigure}[b]{.49\linewidth}
    \includegraphics[width=\textwidth]{figures/Test_bench_Hex_top_frontal.jpg}
    \subcaption{View from $30^ \circ$ angle from the top, showing the manipulator with a red cube in the middle.}
      \end{subfigure}
      \begin{subfigure}[b]{.49\linewidth}
    \includegraphics[width=\textwidth]{figures/Test_bench_side_30_deg.jpg}
    \subcaption{Frontal view without the black shielding, showing the recording from three different camera heights.}
  \end{subfigure}
        \begin{subfigure}[b]{.49\linewidth}
    \includegraphics[width=\textwidth]{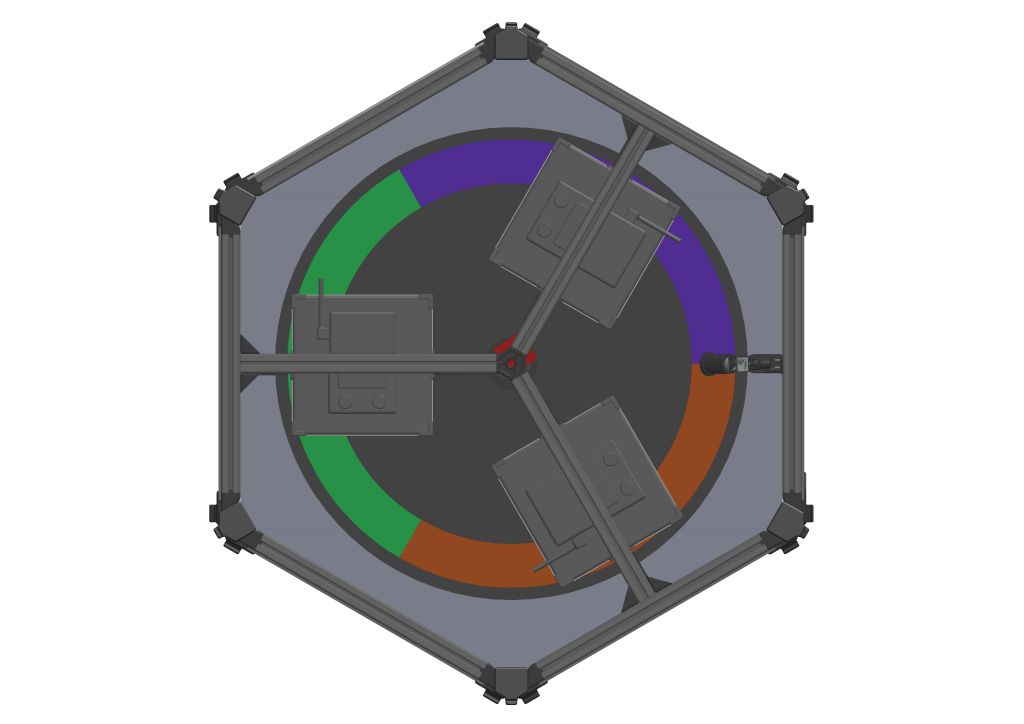}
    \subcaption{View from the top showing the three light centered light panels.}
      \end{subfigure}
      \begin{subfigure}[b]{.49\linewidth}
    \includegraphics[width=\textwidth]{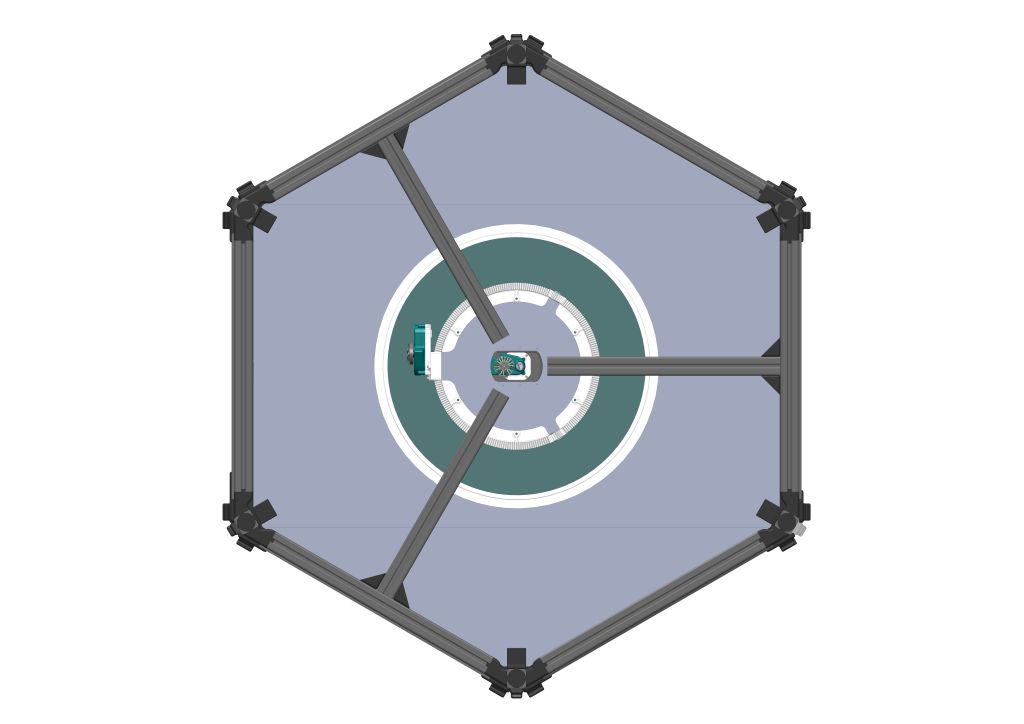}
    \subcaption{View from the bottom, showing the rotating table (turquoise ring) and the manipulator in the center. }
  \end{subfigure}
    \label{fig:renderings_platform_appendix}
    \caption{Different rendered images of the recording setup.}
\end{figure}

\begin{figure}[htp!]
    \centering
    
      \begin{subfigure}[b]{.49\linewidth}
    \includegraphics[width=\textwidth]{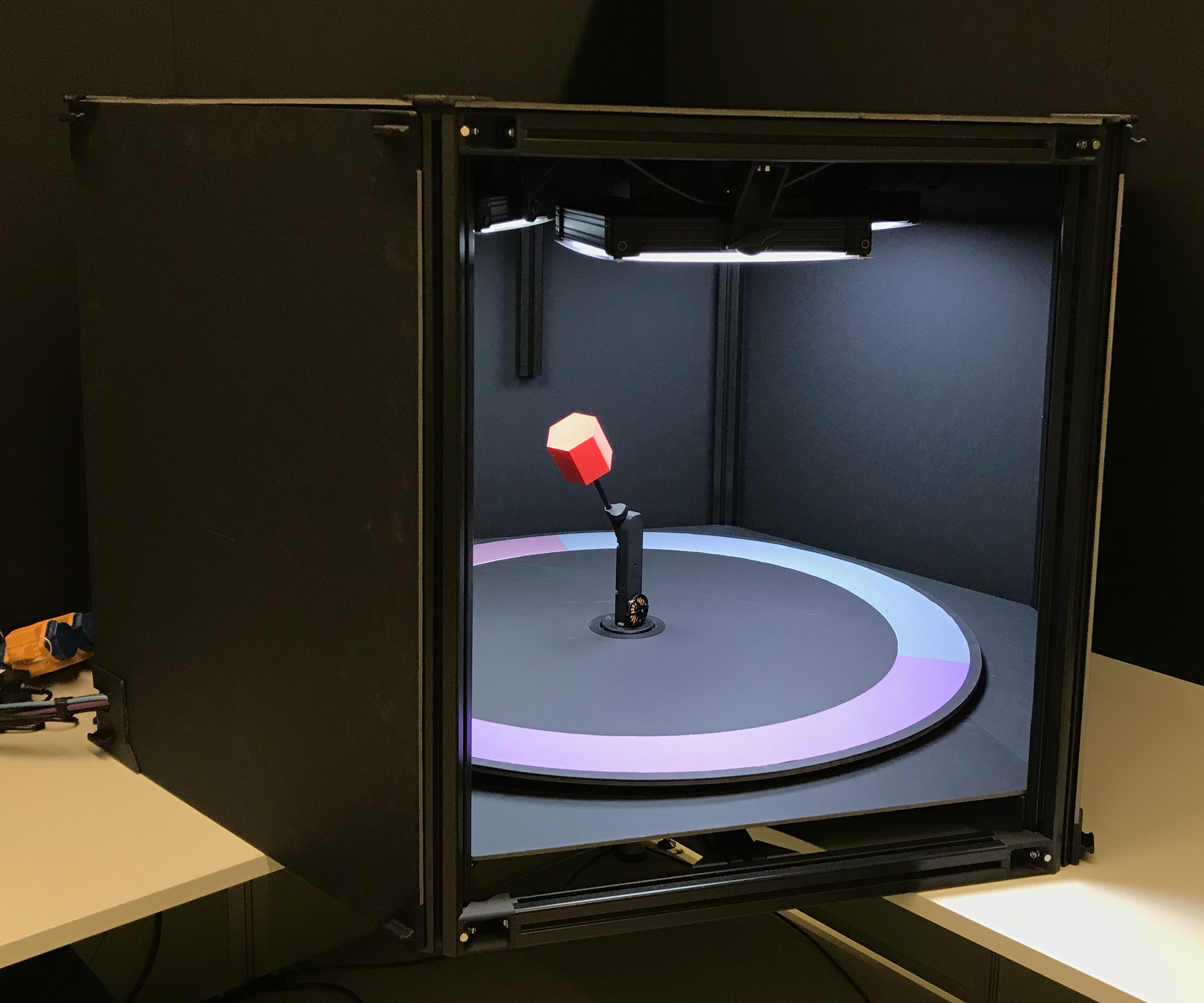}
      \end{subfigure}
      \begin{subfigure}[b]{.49\linewidth}
    \includegraphics[width=\textwidth]{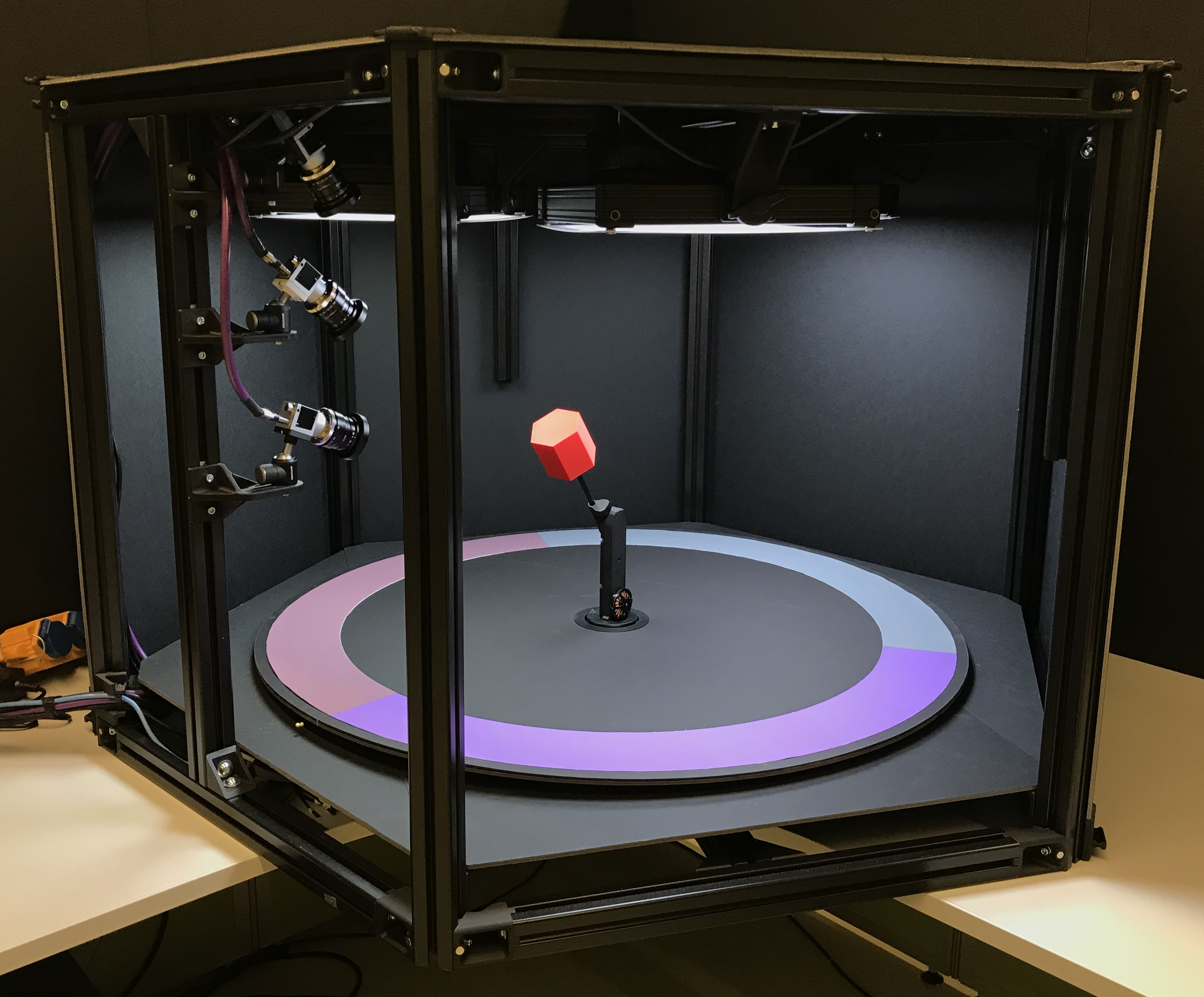}
  \end{subfigure}
    \caption{The mechanical platform for recording the real-world dataset.}
       \label{fig:appendix_platform_real}
\end{figure}

\begin{figure}[htp!]
\centering
    \includegraphics[width=0.5\textwidth]{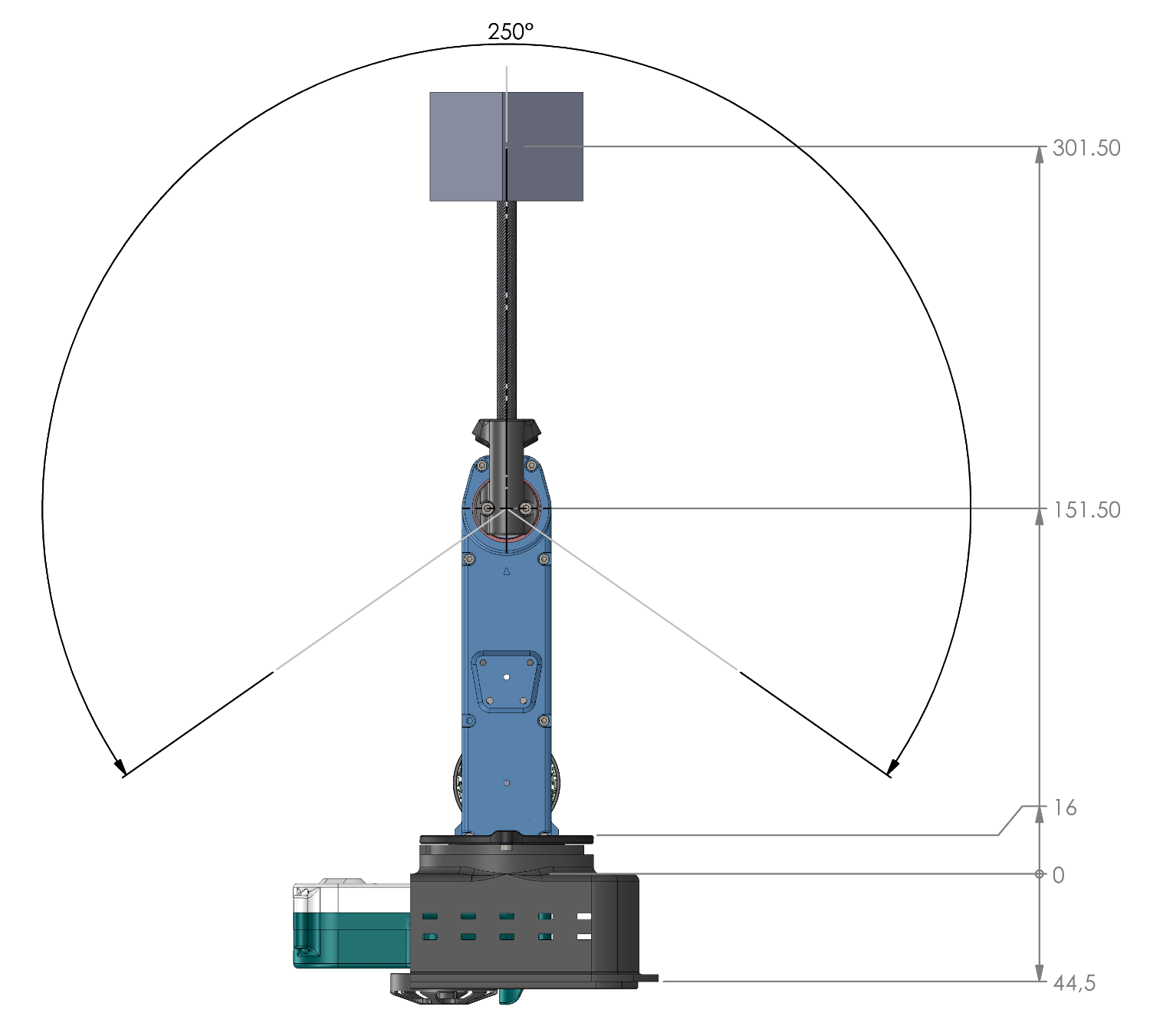}
    \caption{A technical drawing of the manipulator, dimensions are given in mm.}
           \label{fig:technical_drawing}
\end{figure}

\subsection{Difference between Realistic Simulations and Real-World Images}
\label{sec:appendix_difference}

\begin{figure}[htp!]
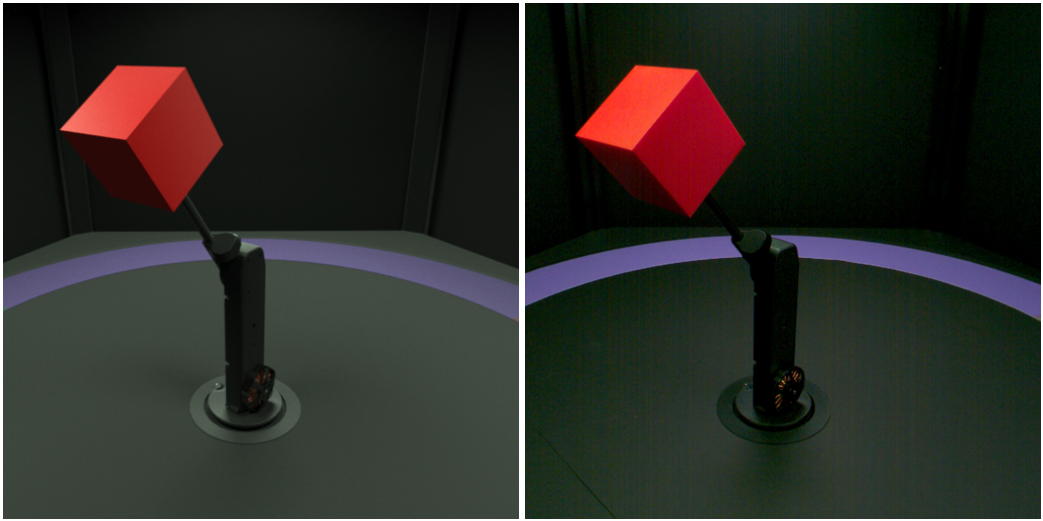

  \centering
  \begin{subfigure}[b]{.49\linewidth}
    \includegraphics[width=\textwidth]{figures/high_resolution_figures/red_cube_realistic.png}
    \subcaption{Realistically simulated sample.}
      \end{subfigure}
      \begin{subfigure}[b]{.49\linewidth}
    \includegraphics[width=\textwidth]{figures/high_resolution_figures/red_cube_real.png}
    \subcaption{Real-world sample.}
  \end{subfigure}
  \caption{Comparison of the realistically simulated and the real-world dataset for one example. The detailed procedure and the availability of CAD files used e.g. for the 3D printed robotic arm ensures a close overlap between both examples. Only small details and a high resolution show the differences e.g. the crack in the floor in the left corner which is only visible in the right image or shadings from the light on the object. Both pictures have a resolution of $512 \times 512$.  }
  \label{fig:high_qualitysimandreal}
\end{figure}

\clearpage
\section{A Discussion on Disentangled Representations and their Transfer}

Empirically, we observed that some of the best trained models were able to disentangle, though imperfectly, the factors of camera height, background colors, object sizes and the motions along the first and second degrees of freedom. Whereas, they performed poorly in disentangling the factors of some object shapes (for e.g. pyramid and cone) and some colors (for e.g. olive and brown). This may well be because of less pixel variation in the respective factors. The features of camera height and background color cause the most difference (maximum L2 distance) in image space. Similarly the object positions at different joint configurations (not the consecutive frames) also have big L2 distance which may explain why the models focus more on learning these factors.

\begin{figure}[!ht]
  \centering
  \begin{subfigure}[b]{\linewidth}
    \includegraphics[width=\textwidth]{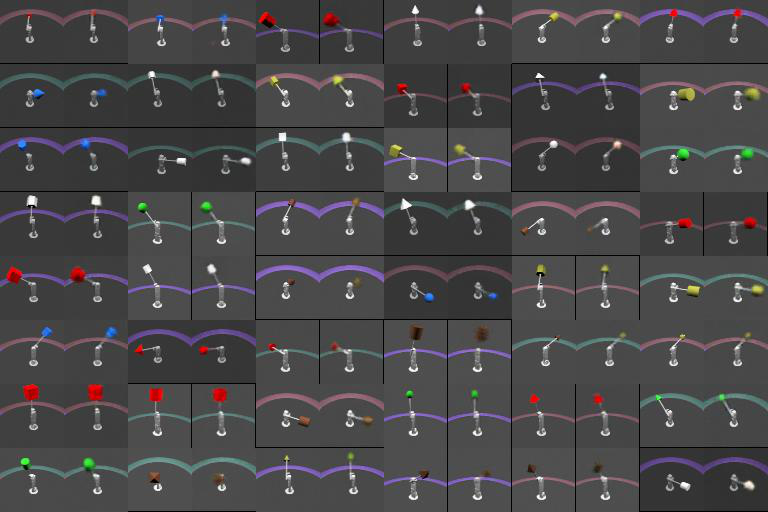}
  \end{subfigure}
  \caption{Image reconstruction of Factor VAE model on the low quality simulated dataset.}
  \label{fig:appendix_simple-reconstruct}
\end{figure}

\begin{figure}[!ht]
  \centering
  \begin{subfigure}[b]{\linewidth}
    \includegraphics[width=\textwidth]{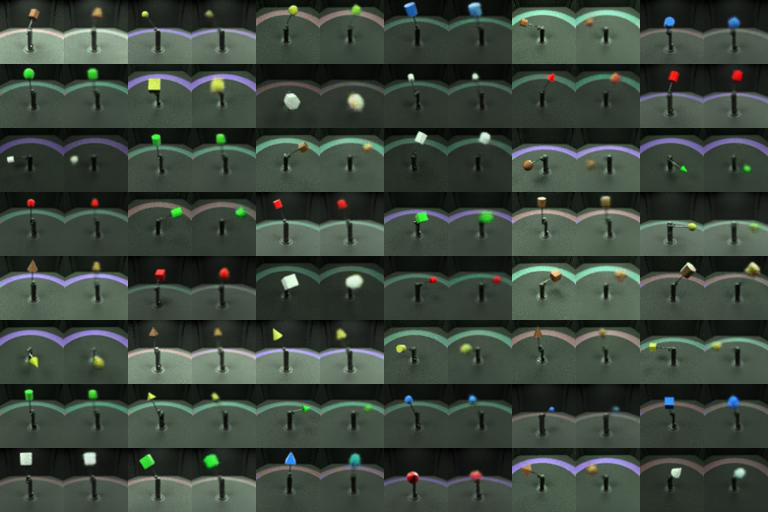}
  \end{subfigure}
  \caption{Image reconstruction of Factor VAE model on the realistic simulated dataset.}
  \label{fig:appendix_realistic-reconstruct}
\end{figure}

\begin{figure}[!ht]
  \centering
  \begin{subfigure}[b]{\linewidth}
    \includegraphics[width=\textwidth]{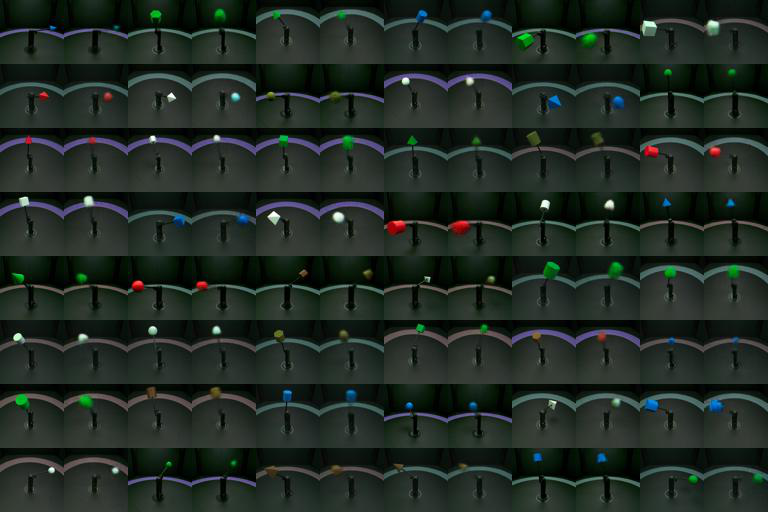}
  \end{subfigure}
  \caption{Image reconstruction of Factor VAE model on the real dataset.}
  \label{fig:appendix_real_reconstruct}
\end{figure}

The image reconstructions become more blurry as we move from the simple simulated dataset to the more complex real-world dataset, which can be seen by the reconstruction scores in figure \ref{fig:reconstruction_complexity}. It has been previously noted that the use of overly simplistic priors like standard normal Gaussians in VAE models \cite{kingma2013auto} can lead to overregularization \cite{tomczak2017vae}. To achieve disentanglement in VAE models, \cite{higgins2017beta} put higher weights ($\beta$>1) on the KL divergence which decreases the reconstruction quality. With the increased complexity in the dataset, the decrease in reconstruction quality becomes even more pronounced, as illustrated in figure \ref{fig:appendix_simple-reconstruct}, \ref{fig:appendix_realistic-reconstruct} and \ref{fig:appendix_real_reconstruct}.

\begin{figure}[!ht]
  \centering
  \begin{subfigure}[b]{\linewidth}
    \includegraphics[width=\textwidth]{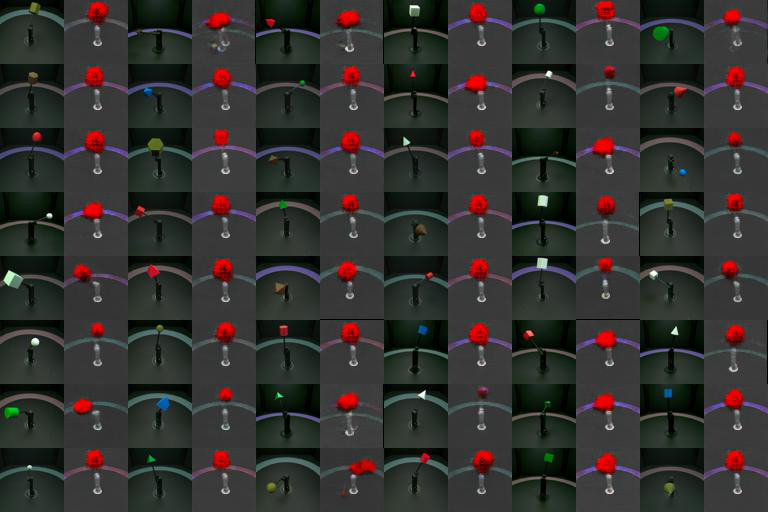}
  \end{subfigure}
  \caption{Image reconstructions of the Factor VAE model trained on toy dataset and tested on the real-world dataset. All images in the uneven columns are real, and to the right of each real image is its reconstruction.}
  \label{fig:appendix_transfer_toy_real_reconstruct}
\end{figure}

\begin{figure}[!ht]
  \centering
  \begin{subfigure}[b]{\linewidth}
    \includegraphics[width=\textwidth]{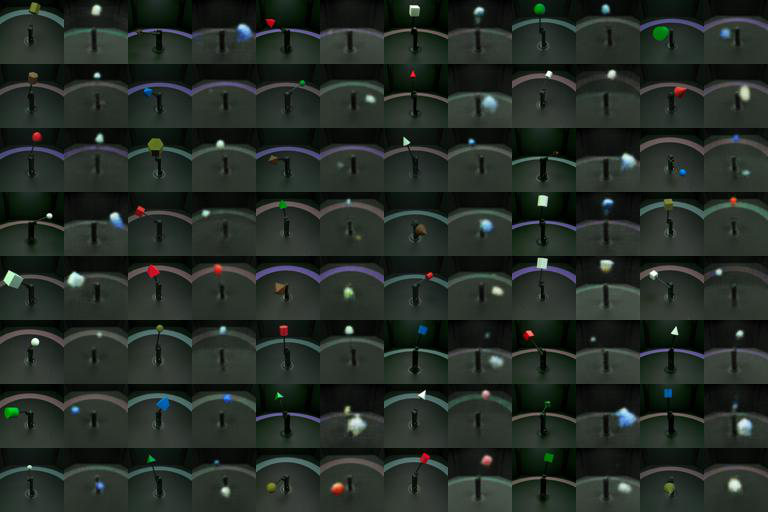}
  \end{subfigure}
  \caption{Image reconstructions of the Factor VAE model trained on realistic simulated dataset and tested on the real-world dataset. All images in the uneven columns are real, and to the right of each real image is its reconstruction.}
  \label{fig:appendix_transfer_realistic_real_reconstruct}
\end{figure}

In figures \ref{fig:appendix_transfer_toy_real_reconstruct} and \ref{fig:appendix_transfer_realistic_real_reconstruct}, we show the reconstruction results of transfers from simple simulated and realistic simulated datasets to the real-world dataset. The models completely fail in transferring the representations from the simple simulated to the real-world data. On the other hand, in the realistic simulated to real-world transfer (figure \ref{fig:appendix_transfer_realistic_real_reconstruct}) the models almost always reconstruct the correct background strip color, manipulator pose and the camera height factor. However, the object properties seem to differ a lot. This shows that in the case of complex environments, the models put more focus on learning the environment to get the better reconstruction accuracy than to learn the important but relatively small changing factors. This result for VAE models has also been confirmed by \cite{ha2018world}.

\section{Details of the Experimental Protocol}
\label{sec:appendix_metrics}

\paragraph{Metrics:} Various methods to validate a learned representation for disentanglement based on known ground truth generative factors $\bm{G}$ have been proposed \cite[e.g.][]{eastwood2018framework,ridgeway2018learning,chen2018isolating, kim2018disentangling}. 
This has for example been expressed as the mutual information of a single latent dimension $Z_i$ with generative factors $G_1,\dots,G_K$ \cite{ridgeway2018learning}, where in the ideal case each $Z_i$ has some mutual information with one generative factor $G_k$ but none with all the others. Similarly, \cite{eastwood2018framework} trained predictors (e.g., Lasso or random forests) for a generative factor $G_k$ based on the representation $\bm{Z}$. In a disentangled model, each dimension $Z_i$ is only useful (i.e., has high feature importance) to predict one of those factors. \cite{suter2018interventional} proposed an interventional robustness score. The graphical model  of \cite{suter2018interventional} adapted to our setup is illustrated in figure \ref{fig:causal-structure}. Another form of validation, especially without known generative factors is the visual inspection of "latent traversals" \cite[see e.g.][]{chen2018isolating}.

\paragraph{Architecture} All the models used the same convolutional encoder and decoder architecture with the fixed latent size of 10.

\begin{table}[htp!]
	\centering
	\begin{tabular}{lcc}
	\toprule
 Model 	& Parameter & Values  \\
\midrule

 $\beta$-VAE & $\beta$ & $[1,\ 2,\ 4,\ 6,\ 8]$\\
 AnnealedVAE & $c_{max}$ & $[5,\ 10,\ 25,\ 50,\ 75]$\\
 & iteration threshold & $100000$\\
 & $\gamma$ & $1000$\\
 FactorVAE & $\gamma$ & $[10,\ 20,\ 30,\ 40,\ 50]$\\
 DIP-VAE-I & $\lambda_{od}$ & $[1,\ 2,\ 5,\ 10,\ 20]$\\
 &$\lambda_{d}$ & $10\lambda_{od}$\\
 DIP-VAE-II & $\lambda_{od}$ & $[1,\ 2,\ 5,\ 10,\ 20]$\\
 &$\lambda_{d}$ & $\lambda_{od}$\\
 $\beta$-TCVAE & $\beta$ & $[1,\ 2,\ 4,\ 6,\ 8]$\\
	\bottomrule
	\end{tabular}
    \caption{Hyperparameters of different methods.}
\label{tab:models}
\end{table}

\begin{table}[htp!]
\centering
\caption{Encoder and Decoder architecture for the main experiment.}
\vspace{2mm}
\begin{tabular}{l  l}
\toprule
\textbf{Encoder} & \textbf{Decoder}\\
\midrule 
Input: $64\times 64 \times$ number of channels & Input: 10\\
$4\times 4$ conv, 32 ReLU, stride 2 & FC, 256 ReLU\\
$4\times 4$ conv, 32 ReLU, stride 2 & FC, $4\times 4\times 64$ ReLU\\
$4\times 4$ conv, 64 ReLU, stride 2 & $4\times 4$ upconv, 64 ReLU, stride 2\\
$4\times 4$ conv, 64 ReLU, stride 2 & $4\times 4$ upconv, 32 ReLU, stride 2\\
FC 256, F2 $2\times 10$ & $4\times 4$ upconv, 32 ReLU, stride 2\\
& $4\times 4$ upconv, number of channels, stride 2\\
\bottomrule
\end{tabular}
\label{table:architecture_main}
\end{table}

\paragraph{Training Hyperparameters} The training hyperparameters were kept fixed for each of the considered methods.
\begin{table}[htp!]
\centering
\caption{Common hyperparameters for training.}
\vspace{2mm}
    \begin{tabular}{l  l}
      \toprule
      \textbf{Parameter} & \textbf{Values}\\
      \midrule 
      Batch size & $64$\\
      Latent space dimension & $10$\\
      Optimizer & Adam\\
      Adam: beta1 & 0.9\\
      Adam: beta2 & 0.999\\
      Adam: epsilon & 1e-8\\
      Adam: learning rate & 0.0001\\
      Decoder type & Bernoulli\\
      Training steps & 300000\\
      \bottomrule
    \end{tabular}
\label{table:fixed_param_main}
\end{table}

\section{Detailed Experimental Results}
\label{sec:appendix_results}

\begin{figure}[htp!] 
 	\centering	
 	\includegraphics[width=\textwidth]{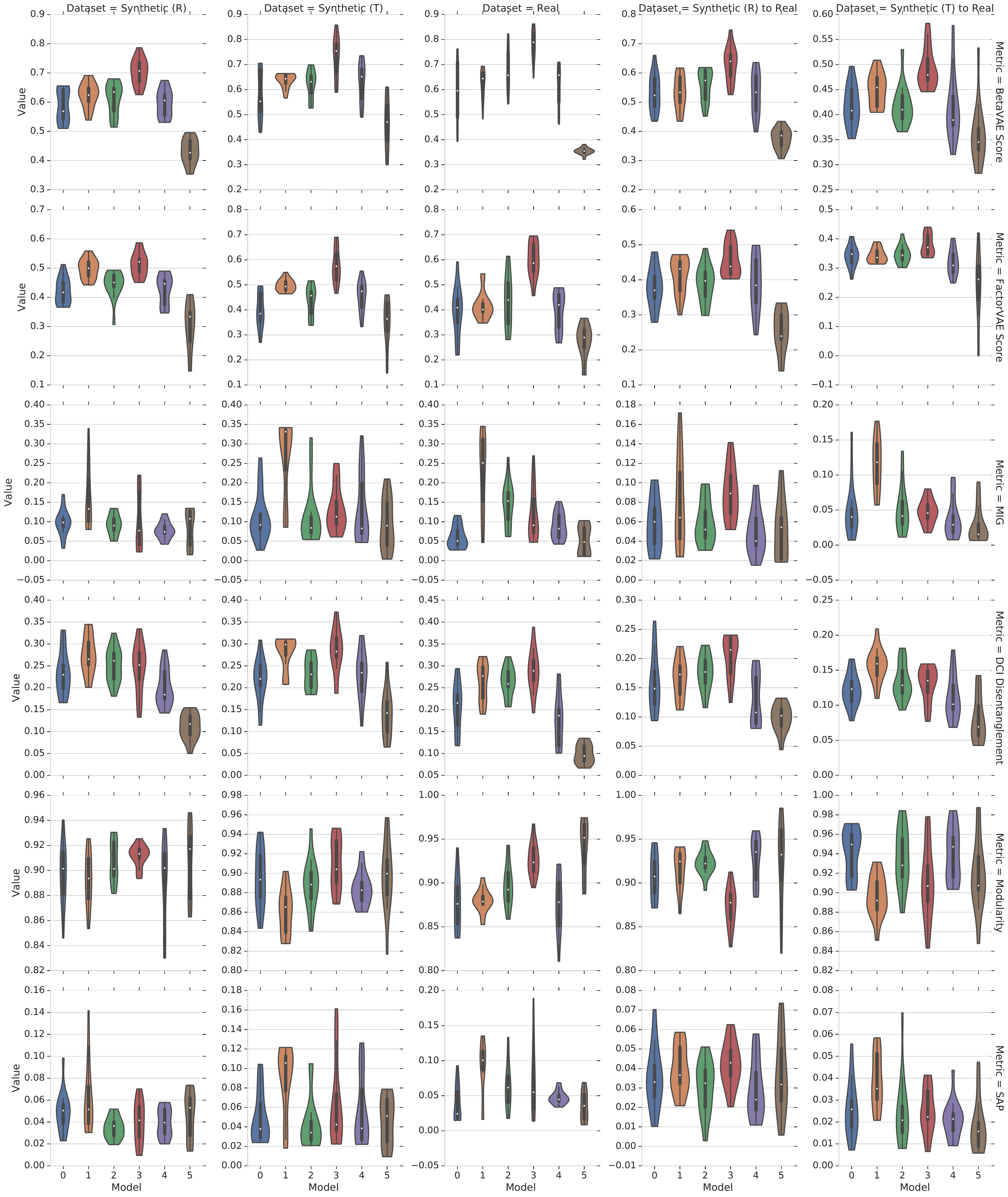}
 	\caption{
 	Disentanglement scores attained by different methods for different metrics (row) and evaluations (column, from left to right): (a) trained and evaluated on synthetic realistic, (b) trained and evaluated on synthetic toy, (c) trained and evaluated on real, (d) trained on synthetic realistic and evaluated on real, (e) trained on synthetic toy and evaluated on real.
 	Methods are abbreviated (0=$\beta$-VAE, 1=FactorVAE, 2=$\beta$-TCVAE, 3=DIP-VAE-I, 4=DIP-VAE-II, 5=AnnealedVAE).
 	}
 	\label{fig:appendix_violin}
 \end{figure}

  \begin{figure}[htp!] 
 	\centering	
 	\includegraphics[width=\textwidth]{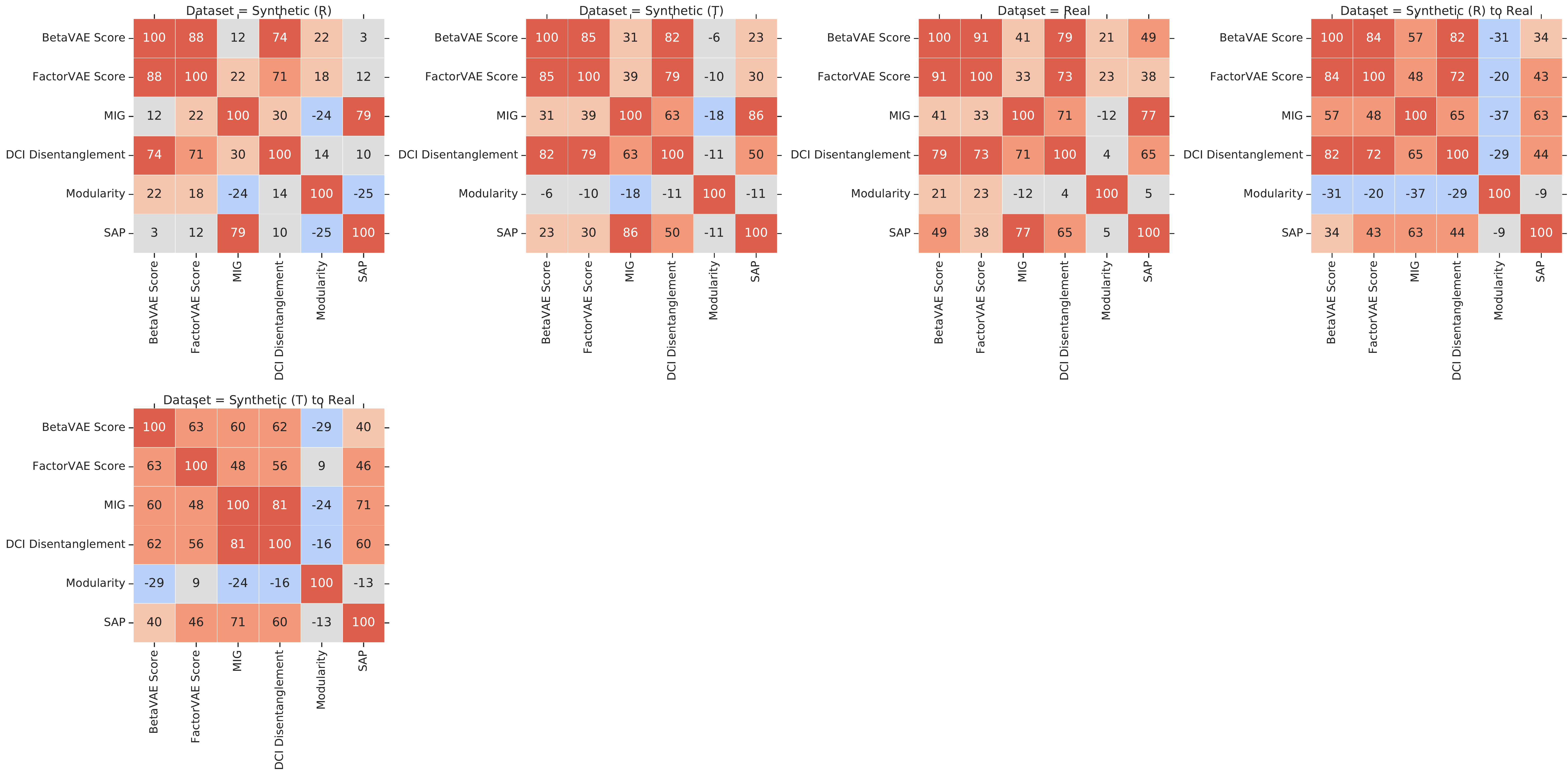}
 	\caption{Rank correlation of different metrics on different data sets. Overall, we observe that all metrics except for  Modularity seem to be at least mildly correlated on all data sets.}
 	\label{fig:rank_correlation}
 \end{figure}
 
   \begin{figure}[htp!] 
 	\centering	
 	\includegraphics[width=\textwidth]{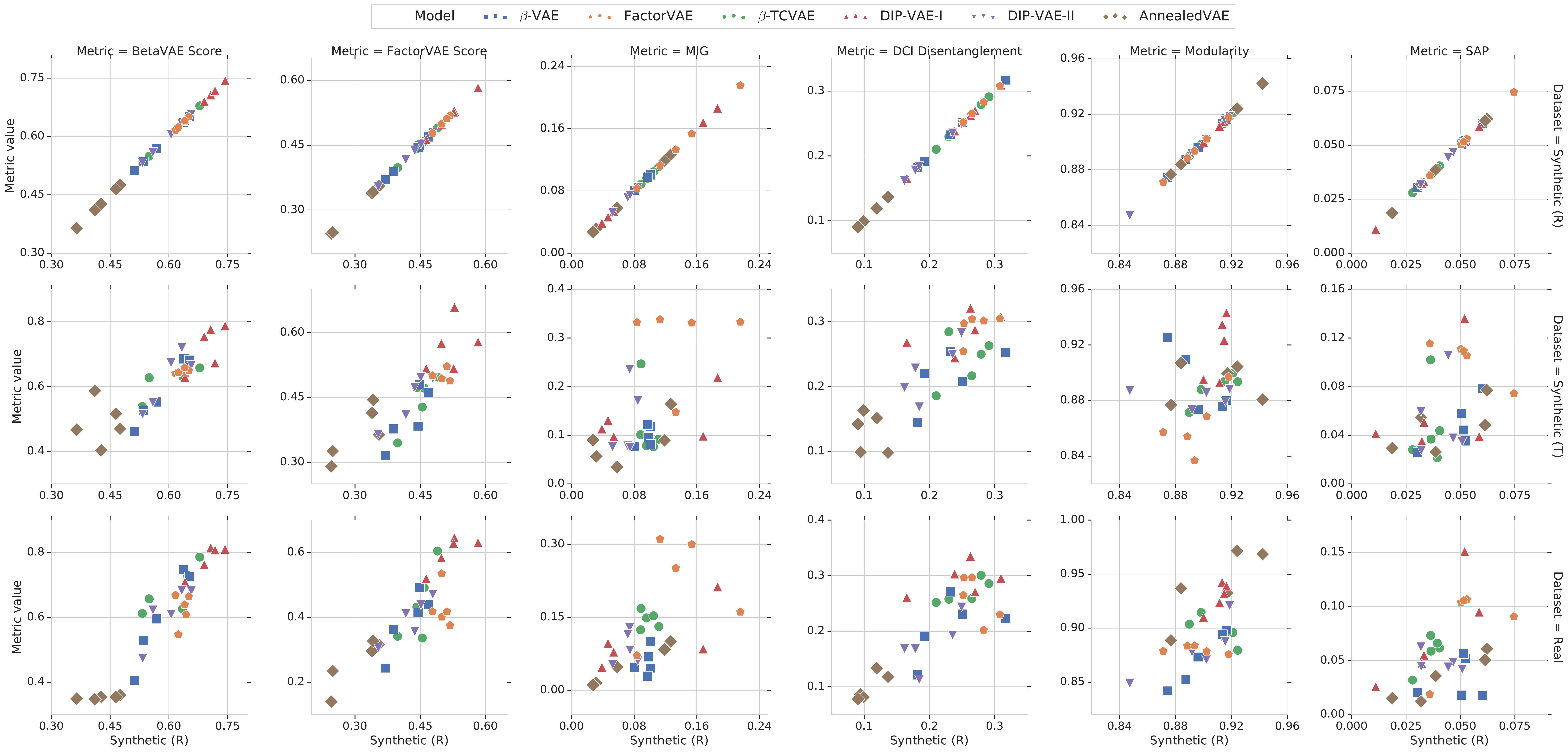}
 	\caption{Disentanglement scores on real data vs simulated datasets. There seems to be a positive correlation for at least three metrics.}
 	\label{fig:scores_realvssimulated}
 \end{figure}

   \begin{figure}[htp!] 
 	\centering	
 	\includegraphics[width=\textwidth]{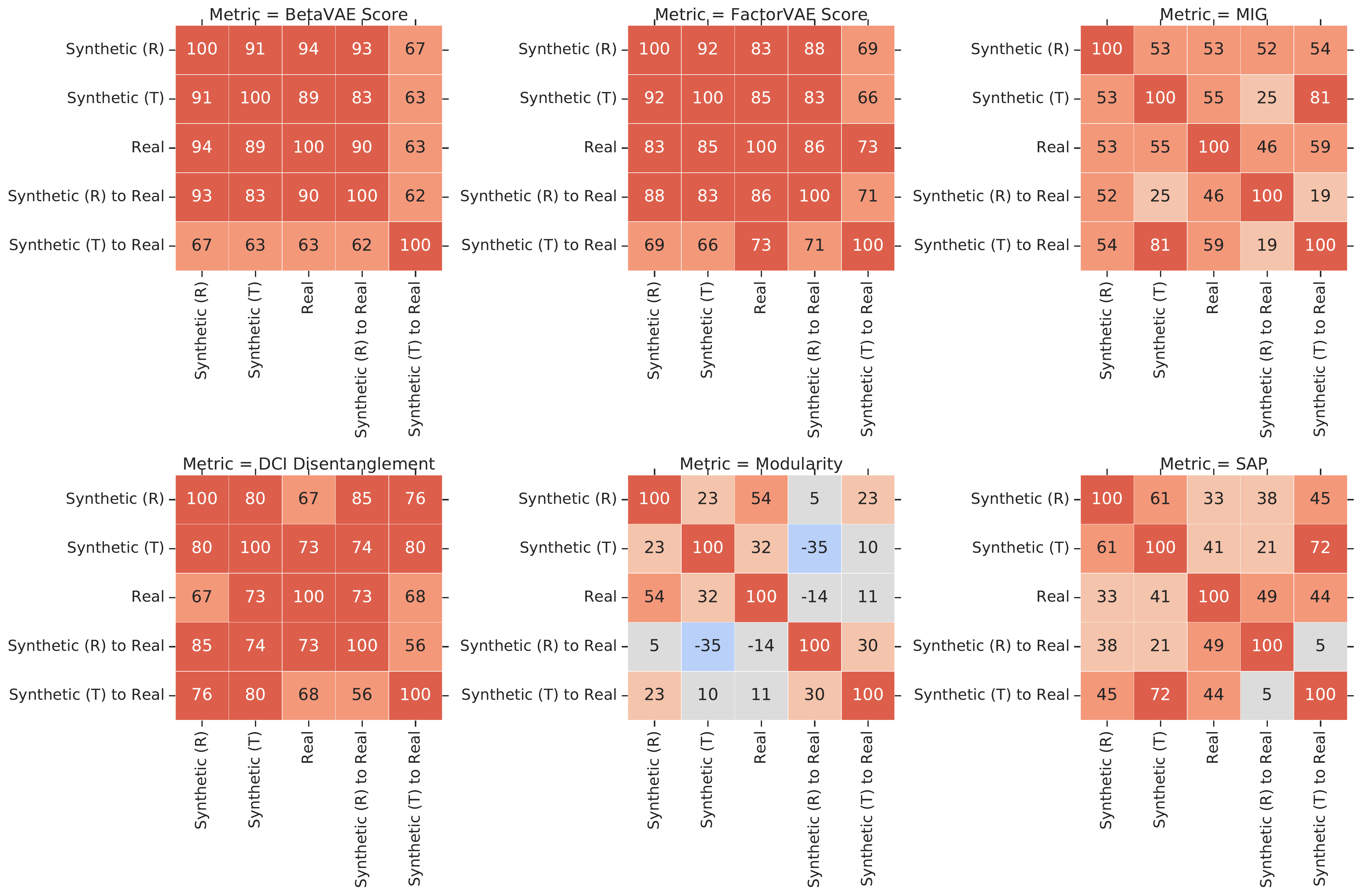}
 	\caption{Rank-correlation of different disentanglement metrics across different data sets. Good hyperparameters seem to transfer well between datasets according to at least three metrics. }
 	\label{fig:rank_across}
 \end{figure}

\end{document}